\documentclass{article}

\usepackage{arxiv}

\usepackage[utf8]{inputenc} 
\usepackage[T1]{fontenc}    
\usepackage{hyperref}       
\usepackage{url}            
\usepackage{booktabs}       
\usepackage{amsfonts}       
\usepackage{nicefrac}       
\usepackage{microtype}      
\usepackage{cleveref}       
\usepackage{lipsum}         
\usepackage{graphicx}
\usepackage{natbib}
\usepackage{doi}

\usepackage{bm}
\let\cite=\citep

\title{Coordinate Encoding on Linear Grids for Physics-Informed Neural Networks}

\date{January 27, 2026}

\newif\ifuniqueAffiliation

\ifuniqueAffiliation 
\author{ \href{https://orcid.org/0000-0000-0000-0000}{\includegraphics[scale=0.06]{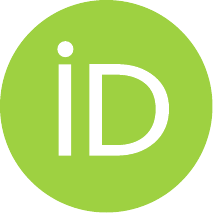}\hspace{1mm}Tetsuro~Tsuchino}\thanks{Use footnote for providing further
		information about author (webpage, alternative
		address)---\emph{not} for acknowledging funding agencies.} \\
	Department of Computer Science\\
	Cranberry-Lemon University\\
	Pittsburgh, PA 15213 \\
	\And
	\href{https://orcid.org/0000-0000-0000-0000}{\includegraphics[scale=0.06]{orcid.pdf}\hspace{1mm}Motoki~Shiga} \\
	Unprecedented-scale Data Analytics Center\\
	Mount-Sheikh University\\
	Santa Narimana, Levand \\
	\texttt{motoki.shiga.b4@tohoku.ac.jp} \\
}
\else
\usepackage{authblk}

\newbox{\orcid}\sbox{\orcid}{\includegraphics[scale=0.06]{orcid.pdf}} 
\author[1,2]{%
	\href{https://orcid.org/0009-0000-1702-1156}{\usebox{\orcid}\hspace{1mm}Tetsuro~Tsuchino}%
}
\author[3,4,5,6]{%
	\href{https://orcid.org/0000-0003-2434-4716}{\usebox{\orcid}\hspace{1mm}Motoki~Shiga\thanks{\texttt{motoki.shiga.b4@tohoku.ac.jp}}}%
}
\affil[1]{Graduate School of Engineering, Gifu University, Japan}
\affil[2]{International Center for Synchrotron Radiation Innovation Smart, Tohoku University, Japan}
\affil[3]{Unprecedented-scale Data Analytics Center, Tohoku University, Japan}
\affil[4]{Graduate School of Information Sciences, Tohoku University, Japan}
\affil[5]{Center for Basic Research on Materials, National Institute for Materials Science, Japan}
\affil[6]{Center for Advanced Intelligence Project, RIKEN, Japan}
\fi

\renewcommand{\shorttitle}{\textit{Coordinate Encoding on Linear Grids for Physics-Informed Neural Networks}}

\hypersetup{
pdftitle={Coordinate Encoding on Linear Grids for Physics-Informed Neural Networks},
pdfsubject={PINNs},
pdfauthor={Tetsuro~Tsuchino, Motoki~Shiga},
pdfkeywords={Physics-Informed Neural Networks, partial differential equations, coordinate encoding, grid-based representation},
}

\begin{document}
\maketitle

\begin{abstract}
In solving partial differential equations (PDEs), machine learning utilizing physical laws has received considerable attention owing to advantages such as mesh-free solutions, unsupervised learning, and feasibility for solving high-dimensional problems. An effective approach is based on physics-informed neural networks (PINNs), which are based on deep neural networks known for their excellent performance in various academic and industrial applications. However, PINNs struggled with model training owing to significantly slow convergence because of a spectral bias problem. In this study, we propose a PINN-based method equipped with a coordinate-encoding layer on linear grid cells. The proposed method improves the training convergence speed by separating the local domains using grid cells. Moreover, it reduces the overall computational cost by using axis-independent linear grid cells. The method also achieves efficient and stable model training by adequately interpolating the encoded coordinates between grid points using natural cubic splines, which guarantees continuous derivative functions of the model computed for the loss functions. The results of numerical experiments demonstrate the effective performance and efficient training convergence speed of the proposed method.
\end{abstract}


\section{Introduction}
Physical laws related to interesting phenomena are often modeled by partial differential equations (PDEs), which include partial derivatives of unknown target functions. Because target functions can rarely be obtained as closed-form solutions, numerical computations are necessary to approximate them. Classical numerical computations are conducted using discrete approximations of continuous functions on predefined spatial grid points, which should be dense for accurate calculation. However, this approach is infeasible for solving high-dimensional PDEs because of the computational burden caused by the large number of grid points in high-dimensional space. 

A machine learning-based approach that utilizes physical laws, known as physics-informed machine learning, has garnered considerable attention over the past decade~\cite{Karniadakis_NatRevPhys2021, Wang_Nature2023, Huang_TNNLS2025}. This approach learns the target functions through regression models whose inputs are system coordinates. The model parameters are trained by minimizing the loss function, which measures the discrepancy between physical laws given by the PDEs and those computed by the model. This approach has two advantages over the classical approaches. The first is a mesh-free solution because the coordinates can be directly inputted into machine learning models. Second, the model training can be conducted in an unsupervised or self-supervised setting, as the approach eliminates the need for target values produced by computationally intensive numerical simulations. An effective machine-learning approach for solving PDEs is based on physics-informed neural networks (PINNs)~\cite{Rassi_JCP2019}, which utilize deep neural networks (DNNs) that have shown excellent performance in various academic and industrial applications~\cite{LeCun_Nature2015, Bishop_DLbook2023deep}. 
Generally, DNN-based models are differentiable with respect to model parameters. This property is utilized for the parameter optimization based on a backpropagation algorithm that leverages efficient gradient computations. The models are also differentiable with respect to model inputs (or system coordinates), which makes it easy to compute the derivatives of the target functions for evaluating the loss functions in the PINN-based approach. These differentials can be automatically computed from any DNN model and inputs using automatic differentiation (AD)~\cite{Baydin_JMLR2018}. The AD technique, which can be implemented using common DNN libraries such as PyTorch, TensorFlow, and JAX, simplifies the implementation of PINNs. PINNs have been applied to various forward and inverse problems in PDEs to understand complex physical phenomena in geoscience and porous media physics~\cite{Kashefi_NNs2023}, biomedical science~\cite{CoaguloNet_NNs2024}, applied physics and medical imaging~\cite{Smyl_NNs2025}, energy science~\cite{RFPINNs_JCP2025}, and industrial applications~\cite{Wang_TNNLS2024, Hua_TNNLS2024}.

The initial PINN model, in which coordinates are fed directly into an MLP to compute target values, incurs high computational costs owing to significantly slow convergence speed during model training. A well-known phenomenon that occurs in training is spectral bias, which prioritizes capturing low-frequency components while failing to capture high-frequency components~~\cite{Rahaman_ICML2019, Wang_JCP2022}. To mitigate spectral bias, coordinate encoding approaches developed in computer vision research ~\cite{Tancik_Neurips2020, K-Planes_CVPR2023} have been utilized in PINNs~\cite{WANG_CMAME2021, PXEL_AAAI2023, Huang_JCP2024}. Effective encoding is based on grid cells, which are used to encode coordinates by interpolating feature vectors on grid points. This technique can mitigate spectral bias by separating all domains into subdomains to extract local high-frequency components. However, these methods have limitations in terms of computational efficiency owing to their grid structures. 

To improve computational efficiency, we propose a new PINN model based on coordinate encoding using linear grid cells defined independently along each axis. This model significantly reduces the number of model parameters and computational cost, particularly for high-dimensional PDEs. For the interpolation of feature vectors, our method incorporates a natural spline to preserve the smoothness of higher-order derivative functions of the model over neighboring grid cells. This incorporation based on the spline is key to stable and efficient training and accurate predictions, which results in its superior performance compared to that of conventional approaches. Our experimental results for benchmark PDE problems demonstrate that the proposed method outperforms competitive methods, such as the initial PINN and methods utilizing grid cells, in terms of training efficiency and prediction performance. 

The remainder of this paper is organized as follows. In Section 2, we review related studies on PINNs. Section 3 outlines our proposed PINN model and its properties. In Section 4, we present comprehensive experimental results for solving PDEs. Finally, Section 5 summarizes our findings and discusses future work.

\section{Related Work}

\subsection{Physics-Informed Neural Network (PINN)}

PINNs are neural-network-based models used for solving PDEs~\cite{Rassi_JCP2019}. In PINNs, model parameters are trained by minimizing the residuals of the governing PDEs on the physical law rather than fitting them to the labeled training data. Recent studies have reported that in the training of PINNs that directly input coordinate vectors, convergence to optimal parameters may be slow and the training iteration may become trapped in local minima~\cite{Krishnapriyan_Neurips2021, Daw_ICML2023, Rathore_ICML2024}. This problem is mainly caused by spectral bias, where the model training prioritizes low-frequency components, and acquiring high-frequency components incurs a huge computational cost~\cite{Rahaman_ICML2019, Wang_JCP2022}. This results in unstable model training and inaccurate predictions.

\subsection{Encoding Coordinates into Feature Vectors}

Encoding coordinates into feature vectors is effective for mitigating spectral bias. One approach is to use basis functions such as the Fourier and Gabor basis functional series. A Fourier-based method, called Fourier (or frequency) encoding~\cite{Tancik_Neurips2020, Shi_2024_CVPR}, transforms the coordinates into the frequency domain and facilitates the extraction of high-frequency components. Alternatively, the basis of Gabor functions, which were originally used in wavelet analysis, is also effective in extracting features localized in the spatial subdomains~\cite{GaborPINNs_GRSL2023}. The effectiveness of PINNs against spectral bias has been demonstrated~\cite{WANG_CMAME2021}. However, it is necessary to select adequate frequency domains for these methods to achieve satisfactory performance.

A flexible approach to solve this problem is based on spatial grid cells in the coordinate axes to separate information among the cells. This approach assumes feature vectors at grid points and optimizes these feature vectors by minimizing the loss functions~\cite{Huang_JCP2024}. Feature vectors for coordinates within cells and not at grids are computed by linearly interpolating the feature vectors at the corners of the cell (grid points). These computed feature vectors are then inputted into neural networks to compute the target values. This approach was first developed in the field of computer vision, and the advantages of high expressive power and the ability to mitigate spectral bias were demonstrated~\cite{K-Planes_CVPR2023}. However, it cannot be applied directly to PINNs because the derivative functions required for the PDE loss function are discontinuous over the grid boundaries, which leads to unstable training. To solve this problem, PIXEL~\cite{PXEL_AAAI2023} uses a nonlinear interpolation with a cosine kernel function to guarantee smoothness for the first-order derivative. Furthermore, it also assumes multiple grids whose cell centers are slightly shifted from the original to reduce the effect of discontinuity over cells in high-order derivatives. Spline-PINN was developed based on Hermite spline kernels, which are piecewise splines in grid cells~\cite{SplinePINNs_AAAI2022}. Hereafter, we refer to Spline-PINN~\cite{SplinePINNs_AAAI2022} as H-Spline. These models demonstrated effective prediction performances for benchmark PDE problems. However, even when using these methods, solving high-dimensional PDEs remains challenging because of the computational cost owing to the exponential order of the number of the grid points.

\subsection{Tensor Factorization Models}

For high-dimensional PDEs, low-rank tensor factorization methods such as canonic-polyadic (CP) and Tucker decompositions and tensor trains have been proposed~\cite{SPINNs_Neurips2023, FunTT_ICPR2025}. These methods first transform each coordinate in the system into a low-rank latent tensor representation, and the latent tensors are then combined using outer products to predict the target values. These methods have shown success in efficiently computing the target values at all the collocation points. However, the spectral bias problem remains unsolved owing to the direct inputs of the coordinates in these models. This problem can be solved by introducing grid cells even for tensor-based models, as demonstrated in our numerical experiments.

\section{Our Proposed Method: Coordinate Encoding on Linear Grid (CELG)}

\subsection{Notations and Problem Setting}

Let $\bm{x}$ be a $D$-dimensional coordinate vector in domain $\Omega \subset \mathcal{R}^D$. The domain is assumed to be a rectangular region that limits the range of the $d$th coordinate axis in $[a_d, b_d]$.  Let $\mathcal{D}[\cdot]$ be the differential operator. A PDE with respect to an unknown function $u(\bm{x}, t)$ is defined as
\begin{eqnarray}
  \label{eq:pde}
  \mathcal{D}[u](\bm{x}, t) = f(\bm{x},t), \quad \bm{x} \in \Omega,~ t\in[0,T]
\end{eqnarray}
where function $f(\bm{x}, t)$ is a source term related to the problem. In addition to the above equation, the initial conditions and boundary conditions
\begin{eqnarray}
  \label{eq:ic_bc}
  u(\bm{x},0) &=& g(\bm{x}), \quad \bm{x} \in \Omega, \\
  \mathcal{B}[u](\bm{x},t) &=& h(\bm{x},t), \quad \bm{x} \in \partial \Omega,~ t\in[0,T].
\end{eqnarray}
are assumed to narrow the solution down to a unique one. $\mathcal{B}[\cdot]$ is the differential operator for the domain boundary $\partial \Omega$. $g(\bm{x})$ and $h(\bm{x},t)$ define the values at the initial state ($t=0)$ and the domain boundary, respectively. 

In a PINN model, the unknown function $u(\bm{x},t)$ is approximated using a neural network with trainable parameters $\bm{\theta}$. Hereafter, we denote the model by $u_{\bm{\theta}}(\bm{x},t)$. Instead of using training datasets, {\it i.e.}, input-output pairs, PINNs use  physical laws described by PDEs to train model parameters. This training task is formulated as a minimization problem for the following loss function: 
\begin{eqnarray}
  l_{\mathrm{total}} (\bm{\theta}) = 
  \lambda_{\mathrm{pde}} \cdot l_{\mathrm{pde}} (\bm{\theta}) 
  + \lambda_{\mathrm{init}} \cdot l_{\mathrm{init}} (\bm{\theta})
  + \lambda_{\mathrm{bc}} \cdot l_{\mathrm{bc}} (\bm{\theta}). 
 \label{eq:loss_total}
\end{eqnarray}
Here, $l_{\mathrm{pde}} (\bm{\theta})$ is a residual function of the governing the PDE describing the physical law and is defined as
\begin{eqnarray}
  l_{\mathrm{pde}} (\bm{\theta}) = \sum_{n=1}^{N_c} \Big( \mathcal{D}[u_{\theta}](\bm{x}_n, t_n) - f(\bm{x}_n, t_n) \Big)^2,
\end{eqnarray}
where $N_c$ denotes the number of collocation points to evaluate the residuals. The determination of these points is described in the section presenting numerical experiments. Evaluating the loss function requires computing the derivative functions of the model with respect to $\bm{x}$ and $t$. The proposed method computes the loss function using automatic differentiation. $l_{\mathrm{init}} (\bm{\theta})$ and $l_{\mathrm{bc}} (\bm{\theta})$ are the loss functions for the initial and boundary conditions, respectively, and are defined as 
\begin{eqnarray}
  l_{\mathrm{init}} (\bm{\theta}) &=&  \sum_{n=1}^{N_i} \Big( u_{\theta}(\bm{x}_n, 0) - g(\bm{x}_n) \Big)^2, \\
  l_{\mathrm{bc}} (\bm{\theta}) &=&  \sum_{n=1}^{N_b} \Big( \mathcal{B}[u_{\theta}](\bm{x}_n, t_n) - h(\bm{x}_n, t_n) \Big)^2,
\end{eqnarray}
where $N_i$ and $N_b$ are the numbers of collocation points used to evaluate the residuals of these conditions. $\lambda_{\mathrm{pde}}$, $\lambda_{\mathrm{init}}$, and $\lambda_{\mathrm{bc}}$ are hyper-parameters that adjust the weights of the three loss functions. 

\subsection{The Proposed Method}

\begin{figure*}[t]
\centering
\includegraphics[width=1\textwidth]{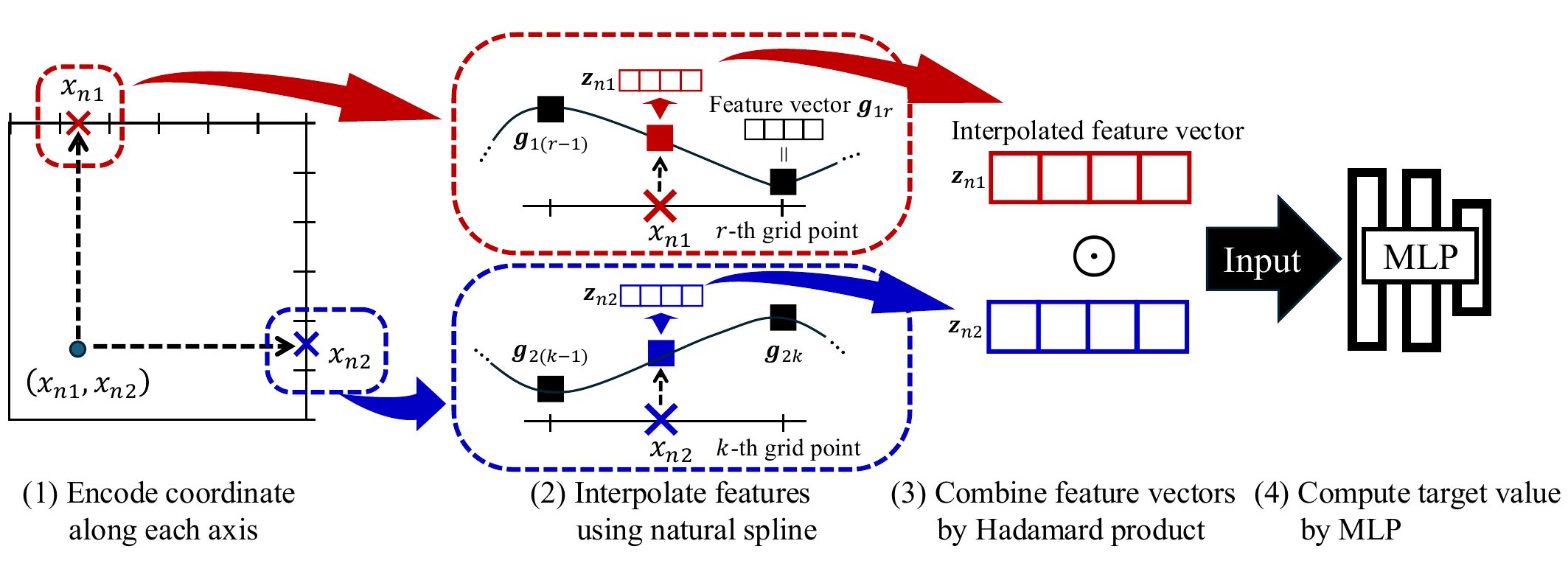}
\caption{Schematic of the proposed method, coordinate encoding on linear grid (CELG).}
\label{fig1_CELG}
\end{figure*}

We propose a PINN model for mitigating spectral bias and improving computational efficiency. The proposed model is illustrated in Fig. \ref{fig1_CELG}. The model first encodes the coordinates (inputs) into feature vectors. Next, the feature vectors are combined using the Hadamard product. The combined vectors are inputted into a multi-layer perceptron (MLP) model to compute the target values. The first procedure is key to improving the prediction performance and computational efficiency, whereas the latter is a common MLP composed of linearly connected layers and nonlinear activation functions. The activation functions are chosen to be higher-order differentiable because the loss function requires higher-order derivatives. Hereafter, we call our proposed method coordinate encoding on linear grid cell (CELG). The following subsections describe the computational procedures in detail.

\subsubsection{Coordinate Encoding}

The coordinates $(\bm{x}, t)$ are first encoded into feature vectors (Fig. \ref{fig1_CELG} (1)--(2)). Encoding is performed by employing feature vectors at the grid points in domain $\Omega$. Our method assumes $R$ grid points independently on each axis. We denote the feature vectors at the grid points as $\bm{g}_{dr}\in\mathbb{R}^M, d=1,\dots,D, r=1,\dots, R$. $\bm{g}$ are also trainable parameters optimized by minimizing the loss function in Eq. (\ref{eq:loss_total}). Our model does not assume feature vectors in the original $D$-dimensional space to reduce the computational complexity, resulting in linear space complexity. However, the existing methods of PIXEL~\cite{PXEL_AAAI2023} and H-Spline~\cite{SplinePINNs_AAAI2022} require exponential spatial complexities because they directly assume feature vectors in the $D$-dimensional space. These computational complexities are discussed in the following section.

The feature vectors for coordinates that are not located at the grid points are computed by interpolating the vectors at the grid points (Fig. \ref{fig1_CELG} (2)). Previous studies have proposed the utilization of a linear interpolation kernel~\cite{PXEL_AAAI2023}, a cosine interpolation kernel by PIXEL~\cite{PXEL_AAAI2023}, and a Hermite spline interpolation kernel by H-Spline~\cite{SplinePINNs_AAAI2022}. Notably, their higher-order derivative functions are discontinuous over grid points during model training, which degrades their training efficiency and prediction performance. The linear interpolation kernel is $C^0$-continuous, whereas the cosine and Hermite interpolation kernels are $C^1$-continuous, as discussed in detail in Appendix \ref{appendix: Hermite-spline}. To avoid this problem, we employ a natural spline for our proposed method because it is always $C^2$-continuous, even during the training. The continuity property of the original and derivative functions over neighboring cells is necessary to perform stable training based on the backpropagation procedure. If this property is violated, the gradients of the PDE loss vanish or blow up during the propagation, resulting in training failure. Even if the training does not fail, an additional training cost is incurred to construct the continuity property satisfied by the PDE solution. Thus, solving this issue is essential for training grid-based approaches, which is discussed in the experimental section by comparing the interpolation methods.

Natural cubic splines are employed because the PDE problems in the numerical experiments require smoothness in the first and second derivatives to evaluate the PDE loss functions. For PDEs with higher-order derivatives, a higher-order natural spline interpolation should be employed.

\subsubsection{Integrating Encoded Features}

The interpolated feature vectors for the coordinates $\bm{z}_{nd} \in\mathbb{R}^M, ~d=1,\dots,D$ are combined using the Hadamard product, 
\begin{eqnarray}
\bm{\phi}_n = \prod_{d=1}^D \bm{z}_{nd},
\label{eq:Hadmard_Prod}
\end{eqnarray}
as shown in Fig. \ref{fig1_CELG} (3). We employ the Hadamard product rather than a simple summation, because multiplicative interactions between axis-wise features provide more accurate function approximations, as proven theoretically~\cite{Jayakumar_ICLR2020} and demonstrated numerically~\cite{K-Planes_CVPR2023}.

The combined vectors $\bm{\phi}_n, ~n=1,\dots,N$ are used to compute the target values $u_{\theta}(\bm{x}_n), ~n=1,\dots,N$ (Fig. \ref{fig1_CELG} (4)). For the computation, we employed an MLP to capture nonlinear interactions across the coordinate axes in the system. For the activation functions in the MLP, we chose nonlinear and higher-order differentiable functions to avoid the problem of computing the loss functions for PINNs and to capture the nonlinear interaction. We employed the $\tanh$ function as the activation function.

The proposed model appears similar to existing tensor-based models~\cite{SPINNs_Neurips2023, FunTT_ICPR2025}. These existing models first separately approximate functions for each axis using MLPs and then combine them by utilizing tensor factorization, {\it i.e.}, linear combinations. It has been theoretically proven that these existing models can approximate any continuous function within a small error. However, these existing models suffer from two problems: spectral bias and high computational costs. The former is caused by MLPs, which require large training steps to capture high-frequency components, even for a one-dimensional function. Our model avoids this problem by using coordinate encoding. The latter is caused by the tensor factorization, which captures the nonlinear interactions across the coordinate axes. The tensor-based approach is advantageous for reducing the number of collocation points for training, thereby accelerating the computational speed. However, capturing these interactions requires a sufficiently high tensor rank. This leads to a significant increase in the number of model parameters, resulting in high computational costs. Consequently, the existing models are intractable for high-dimensional nonlinear problems. Instead of tensor factorization, our model employs an MLP to capture interactions. For a numerical comparison, we implemented another model in which the MLP in CELG was replaced with CP factorization. Hereafter, this model is referred to as CELG-CP.

\subsection{Computational Complexity of Coordinate Encoding}
\label{sec:complexity}

This section discusses the computational complexities of coordinate encoding procedures used in the proposed method (CELG) and competitive methods (H-Spline~\cite{SplinePINNs_AAAI2022} and PIXEL~\cite{PXEL_AAAI2023}). The space complexity of CELG is $\mathcal{O}(M\cdot D\cdot R)$ because it has $M$-dimensional feature vectors at $R$ grid points along each $D$ axes. In contrast, H-Spline and PIXEL assume grid cells in the original $D$-dimensional space, generating grid points at a combination of $R$ points along $D$ axis. H-Spline employs weights in the Hermite interpolation kernel functions at the grid points for the interpolation. For an $S$-th order Hermite spline, the length of each coefficient vector is $(S+1)^D$. Thus, the space complexity of Hermite spline interpolation is $\mathcal{O}(S^D \cdot R^D)$. PIXEL has $M$-dimensional feature vectors at the grid points, and coordinate encoding is performed by interpolating these vectors using a cosine-based interpolation kernel. Therefore, the space complexity is $\mathcal{O}(M\cdot R^D)$. H-Spline and PIXEL require a space complexity of an exponential order with respect to the dimension, which is infeasible for high-dimensional PDEs.

The time complexity of CELG for computing a coordinate encoding is linear because the linear equations used to determine the natural cubic spline coefficients can be solved efficiently using a tridiagonal matrix algorithm whose order is $\mathcal{O}(R)$. Therefore, the time complexity is $\mathcal{O}(M\cdot D\cdot R)$. However, the time complexities of H-Spline and PIXEL are $\mathcal{O}(S^D\cdot 2^D)$ and $\mathcal{O}(M\cdot 2^D)$, respectively, because these interpolations use feature vectors at the corners of the grid cell that includes the target coordinate, whose number is $2^D$. Overall, CELG is superior to these competitive methods in terms of both space and time complexities. This improvement is important because training PINN models requires a large number of epochs, particularly for high-dimensional PDEs. The time and space complexities are summarized in Table \ref{table1_Complexity}.

\begin{table}[tb]
  \caption{Computational complexities}
  \label{table1_Complexity}
  \small
  \begin{center}
  \begin{tabular}{cccc}
    \hline
    & H-Spline & PIXEL & CELG \\
    \hline
    Space &
    $\mathcal{O}(S^D\cdot R^D)$ & $\mathcal{O}(M\cdot R^D)$ & $\mathcal{O}(M\cdot D\cdot R)$ \\
    Time &
    $\mathcal{O}(S^D\cdot 2^D)$ & $\mathcal{O}(M\cdot 2^D)$ & $\mathcal{O}(M\cdot D\cdot R)$ \\
    \hline
  \end{tabular}
  \end{center}
\end{table}

\section{Numerical Experiments}

This section describes the numerical experiments performed using the proposed method and competitive methods, providing a comparison of the results. All experiments were implemented on a computer with an Intel Xeon Gold 6230 CPU, with 768 GB of memory and an NVIDIA A100-PCIE-40GB GPU. The code to implement all the models and automatic differentiation was created using Python 3.10.12 and JAX version 0.5.0. All the model parameters were optimized using a backpropagation algorithm adapted from Adam~\cite{Kingma_Adam_ICLR2015}. Our developed code is available in the GitHub repository (the URL will be added after acceptance).

\subsection{PDEs}

This section describes the PDEs computed in the experiments.

\subsubsection{Multi-Band Poisson Equation}

A one-dimensional multi-band Poisson equation was defined in a previous study~\cite{Deepxde_SIAMRev2021}. We generalized it to a multi-dimensional equation as follows:
\begin{eqnarray}
 \Delta u(\bm{x}) &=&  - \sum_{i=1}^I \sum_{d=1}^D w_i \pi^2 c_i^2 \sin(\pi c_i x_d),~ \bm{x} \in \Omega \\
 u(\bm{x}) &=& u_{\mathrm{true}}(\bm{x}),~~ \bm{x} \in \partial \Omega,
\end{eqnarray}
where the number of bands $I$, weight $w_i$, and frequency $c_i$ are predefined system parameters.
$\Delta$ is Laplacian operator defined by 
\begin{eqnarray}
  \Delta u(\bm{x}) = \sum_{d=1}^{D} \frac{\partial^2}{\partial x_d^2} u(\bm{x}).
\end{eqnarray}
Under the assumption of the additive functional form with respect to each coordinate axis, the analytical solution is obtained as follows:
\begin{eqnarray}
 u_{\mathrm{true}}(\bm{x}) =  \sum_{i=1}^I \sum_{d=1}^D w_i \sin{(\pi c_i x_d)}.
\end{eqnarray}

\subsubsection{Burgers Equation}
The equation is defined as follows:
\begin{eqnarray}
\frac{\partial}{\partial t} u(x,t) + u(x,t)\cdot \frac{\partial}{\partial x} u(x,t) - \nu\frac{\partial^2}{\partial x^2} u(x,t)=0,
\nonumber \\
~x\in[-1,1],~t\in[0,1].
\end{eqnarray}
 The initial and boundary conditions are given by 
\begin{eqnarray}
u(x,t) &=& -\sin(\pi x), \\
u(-1,t) &=& u(1,t) = 0.
\end{eqnarray}
In our experiments, the system parameter was fixed at $\nu=0.01/\pi$.
Because the solution cannot be obtained analytically, we used a precise numerical solution provided in a previous study~\cite{Rassi_JCP2019} as the ground truth.

\subsubsection{Allen-Cahn Equation}
The equation is defined as follows:
\begin{eqnarray}
\frac{\partial}{\partial t} u(x,t) - \nu \frac{\partial^2}{\partial x^2} u(x,t) + 5 u^3(x,t) - 5 u(x,t) = 0,
\nonumber \\
~x\in[-1,1],~t\in[0,1].
\end{eqnarray}
The initial and boundary conditions are given by
\begin{eqnarray}
u(x,0) &=& x^2 \cos(\pi x), \\
u(-1,t) &=& u(1,t), \\
\frac{\partial}{\partial x} u(-1,t) &=& \frac{\partial}{\partial x} u(1,t).
\end{eqnarray}
In our experiments, the system parameter was fixed at $\nu=0.0001$.
Because the solution cannot be obtained analytically, we used a precise numerical solution provided in a previous study~\cite{Rassi_JCP2019} as the ground truth.

\subsubsection{Flow-Mixing Equation}
The equation is given by
\begin{eqnarray}
 \frac{\partial}{\partial t} u(\bm{x},t) + a(x_1,x_2)\frac{\partial}{\partial x_1} u(\bm{x},t) && 
\nonumber \\
 + b(x_1,x_2)\frac{\partial}{\partial x_2} u(\bm{x},t)&=& 0,
\nonumber \\
~\bm{x}\in\Omega,~t\in[0,8],
\end{eqnarray}
where
\begin{eqnarray}
a(x_1, x_2) &=& -\frac{v_t}{v_{\mathrm{tmax}}}\frac{x_2}{r} \\
b(x_1, x_2) &=& \frac{v_t}{v_{\mathrm{tmax}}}\frac{x_1}{r} \\
v_t &=& \mathrm{sech}^2(r) \tanh(r) \\
r &=& \sqrt{x_1^2 + x_2^2}.
\end{eqnarray}
The initial and boundary conditions are given by 
\begin{eqnarray}
u(\bm{x},0) &=& u_{\mathrm{true}}(\bm{x},0),~~ \bm{x}\in\Omega, \\
u(\bm{x}, t) &=& u_{\mathrm{true}}(\bm{x},t),~~ \bm{x} \in \partial \Omega,
\end{eqnarray}
where the domain is $\Omega={[-4,4]}^2$. The analytical solution is obtained as follows:
\begin{eqnarray}
  u_{\mathrm{true}}(\bm{x},t) = -\tanh\left( \frac{x_2}{2}\cos(\omega t) - \frac{x_1}{2}\sin(\omega t) \right),
\end{eqnarray}
where $\omega=\frac{1}{r}\frac{v_t}{v_{\mathrm{tmax}}}$. The system parameter was fixed at $v_{\mathrm{tmax}} = 0.385$ in all the experiments.

\subsubsection{Helmholtz Equation}
The equation is defined as follows:
\begin{eqnarray}
\Delta u(\bm{x}) + k^2 u(\bm{x}) = -k^2(D-1) \prod_{d=1}^{D} \sin(k x_d),~ \bm{x} \in \Omega \\
u(\bm{x}) = 0, \quad \bm{x} \in \partial \Omega,
\end{eqnarray}
where the domain $\Omega={[0,1]}^D$ and $k$ is the parameter about the wave number. 
The analytical solution is obtained as follows:
\begin{eqnarray}
  u_{\mathrm{true}}(\bm{x}) = \prod_{d=1}^{D} \sin(k x_d).
\end{eqnarray}
The wavenumber parameter was fixed at $k=10\pi$ in all the experiments.

\begin{figure}[t]
\centering
\includegraphics[width=0.7\columnwidth]{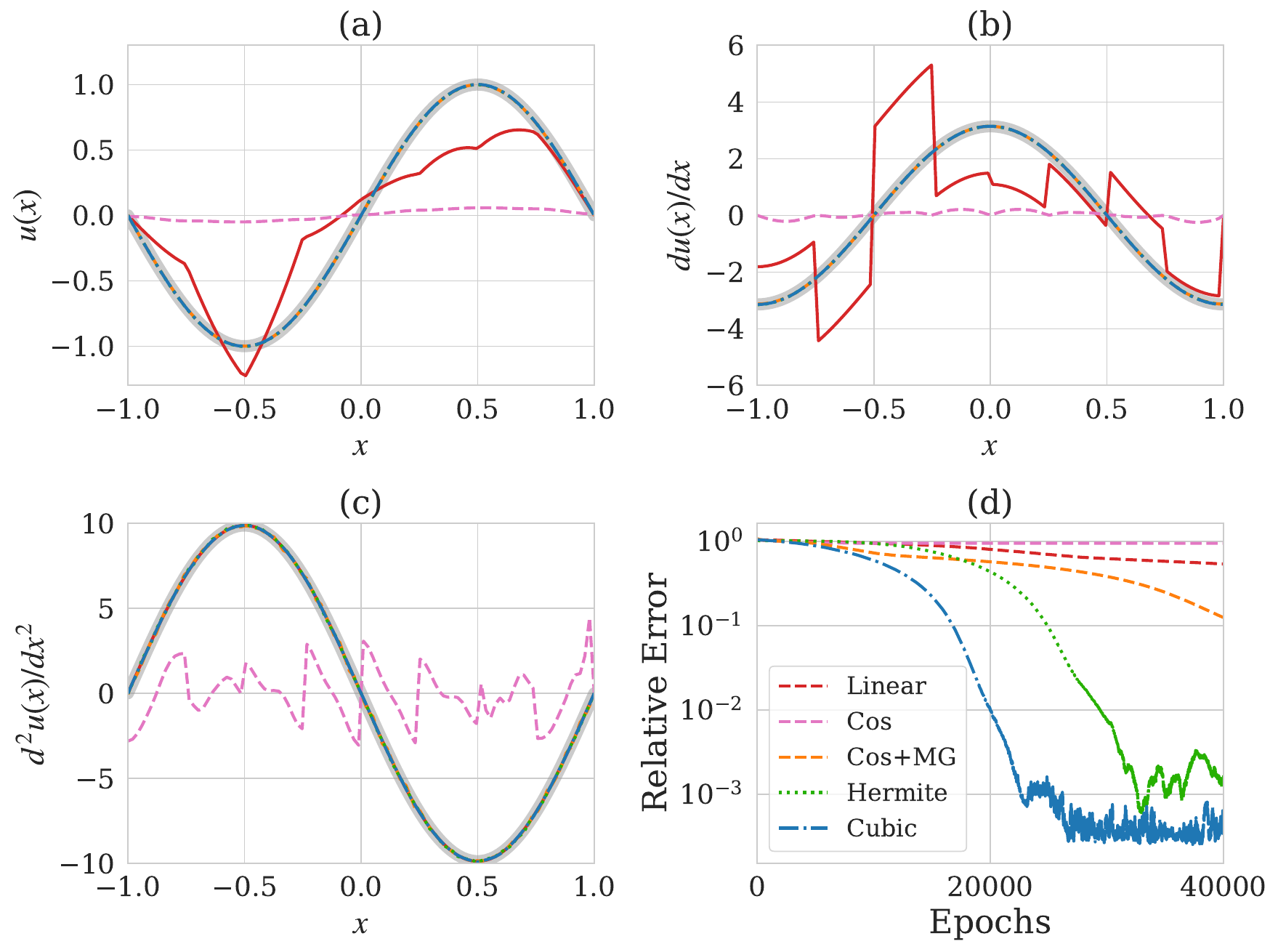} 
\caption{Training results utilizing linear, cosine with/without multigrid (MG), Hermite and natural interpolations for 1D Poisson PDE. (a) Predictions of $u(x)$, (b) first and (c) second derivatives. (d) Learning curves. In (a)--(c), the analytical solution is shown as a solid gray line.}
\label{fig2_interp_result}
\end{figure}

\subsection{Comparison of Interpolation Methods}

In a grid-based approach, the feature vector interpolation is crucial for efficient training and accurate predictions. This section discusses the validity of the proposed method based on natural cubic splines by comparing it with conventional interpolation methods. The compared methods were linear interpolation, the cosine-based method used in PIXEL, and the Hermite spline used in H-Spline. Linear interpolation is continuously differentiable, {\it i.e.,} $C^1$, resulting in zero higher-order derivatives. The cosine-based method is infinitely differentiable; however, its second- and higher-order derivatives are not smooth across cell boundaries. The Hermite spline interpolation is a higher-order differentiability through the appropriate choice of kernel functions and is smooth across cell boundaries. However, the narrow support of its kernel functions limits the propagation of encoded features and their gradients to only immediate neighboring cells during training and prediction. Appendix A discusses the properties of the cosine and Hermite kernel functions in detail.
We applied these methods to the single-band 1D Poisson PDE with parameters $w_1=1.0$ and $c_1=1$. The PDE requires second-order derivatives to compute the loss function. For these methods, the grid cell interval was set to 0.25, {\it i.e.,} $R=8$, in the domain $\Omega=[-1,+1]$. The loss weights were set to $\lambda_{\mathrm{pde}}=1$ and $\lambda_{\mathrm{bc}}=0.01$. One hundred and twenty-eight collocation and test points were placed at the regular interval over the domain. The learning ratio of these methods was set to $10^{-6}$.

Figs. \ref{fig2_interp_result} (a)-(c) show the results predicted by our grid cell-based models with four interpolations. These figures show that the non-smoothness of the derivatives of the linear and cosine interpolations severely impaired the prediction of $u(x)$ and its first-order derivatives, although they performed well in fitting the second-order derivative functions. In contrast, the natural cubic and Hermite splines, both of which satisfy the smoothness of their derivatives, correctly predicted all the functions. To mitigate the non-smooth property of cosine interpolation, a multigrid (MG) approach was proposed to place overlapping cell regions across the cell boundaries~\cite{PXEL_AAAI2023}. Fig. \ref{fig2_interp_result} shows that the method (Cos+MG) yielded accurate predictions. 

Fig. \ref{fig2_interp_result} (d) shows the learning curves of the models with these interpolations. The relative errors were computed using the mean-squared errors between the ground truth and predictions, normalized by the $\ell_2$ norm of the ground truth. This figure shows that training using Hermite and cosine-based interpolations is slower than that using natural cubic interpolation. 
This behavior can be explained by how the interpolation kernels propagate information. The cubic spline enforces $C^2$ continuity across cell boundaries, allowing the feature and gradient information to propagate smoothly across cells, whereas Hermite interpolation allows feature-vector information to propagate only to the neighboring cells owing to its localized kernel functions. Furthermore, the cosine-based method (Cos+MG) exhibited a much slower convergence than the other methods owing to the overlapped non-smooth grid cells. Overall, the natural cubic spline interpolation was the most suitable method for interpolating feature vectors in our setting.

\begin{figure}[t]
\centering
\includegraphics[width=0.7\columnwidth]{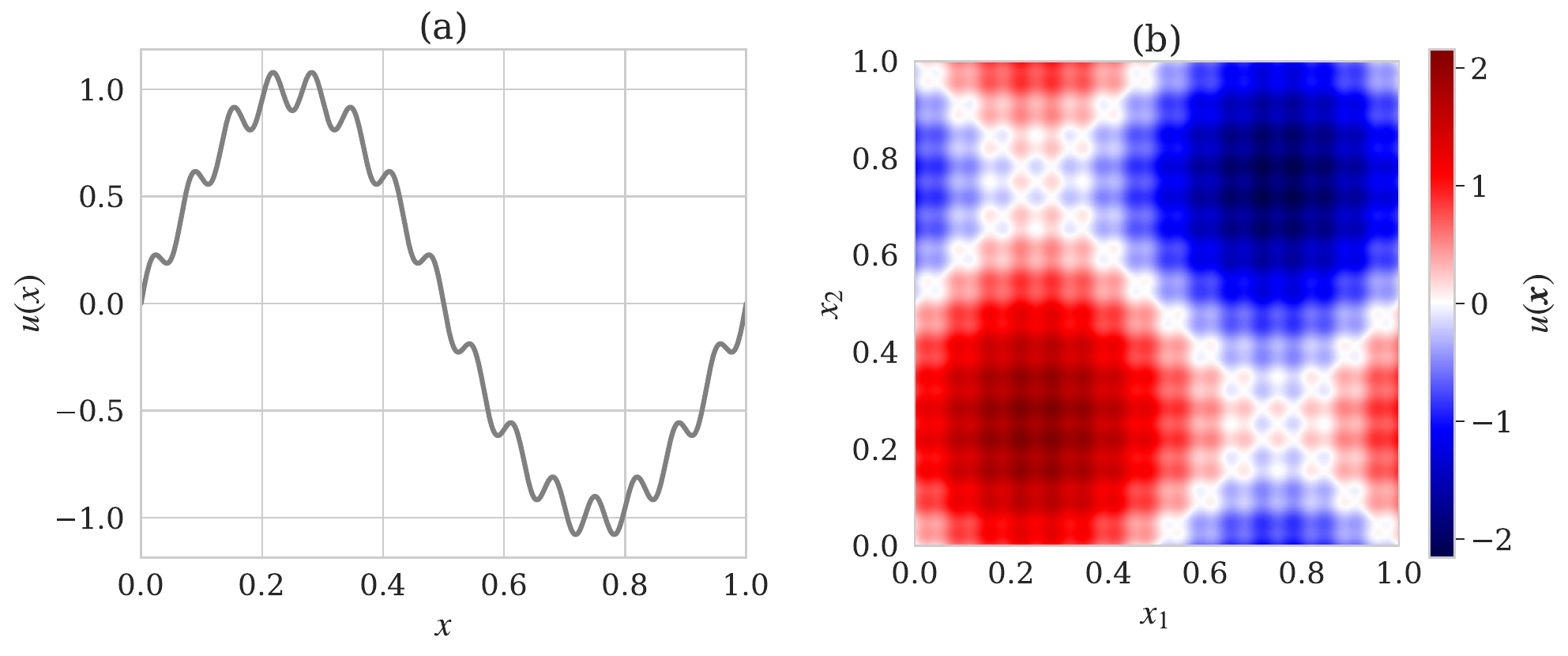} 
\caption{Analytical solutions of (a) 1D and (b) 2D multi-band Poisson PDEs.}
\label{fig3_PissonTrue}
\end{figure}

\begin{figure}[t]
\centering
\includegraphics[width=0.7\columnwidth]{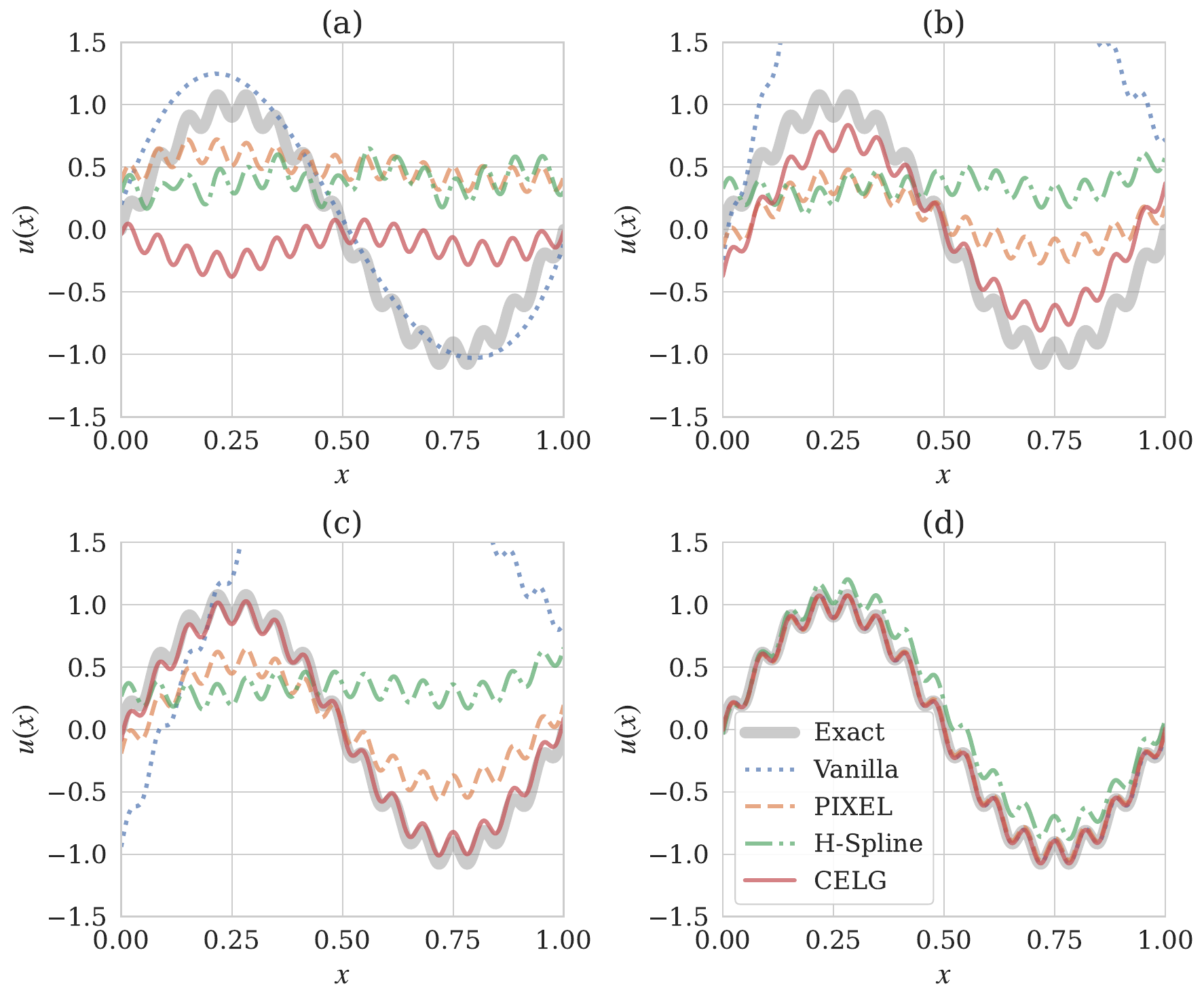} 
\caption{Prediction results for the 1D multi-band Poisson PDE during training at (a) 100, (b) 1,000, (c) 5,000, and (d) 10,000 training epochs. The grid parameter was set to $R = 16$ in PIXEL, H-Spline, and CELG.}
\label{fig4_Pisson1D_ResultTrue}
\end{figure}

\begin{figure}[t]
\centering
\includegraphics[width=0.7\columnwidth]{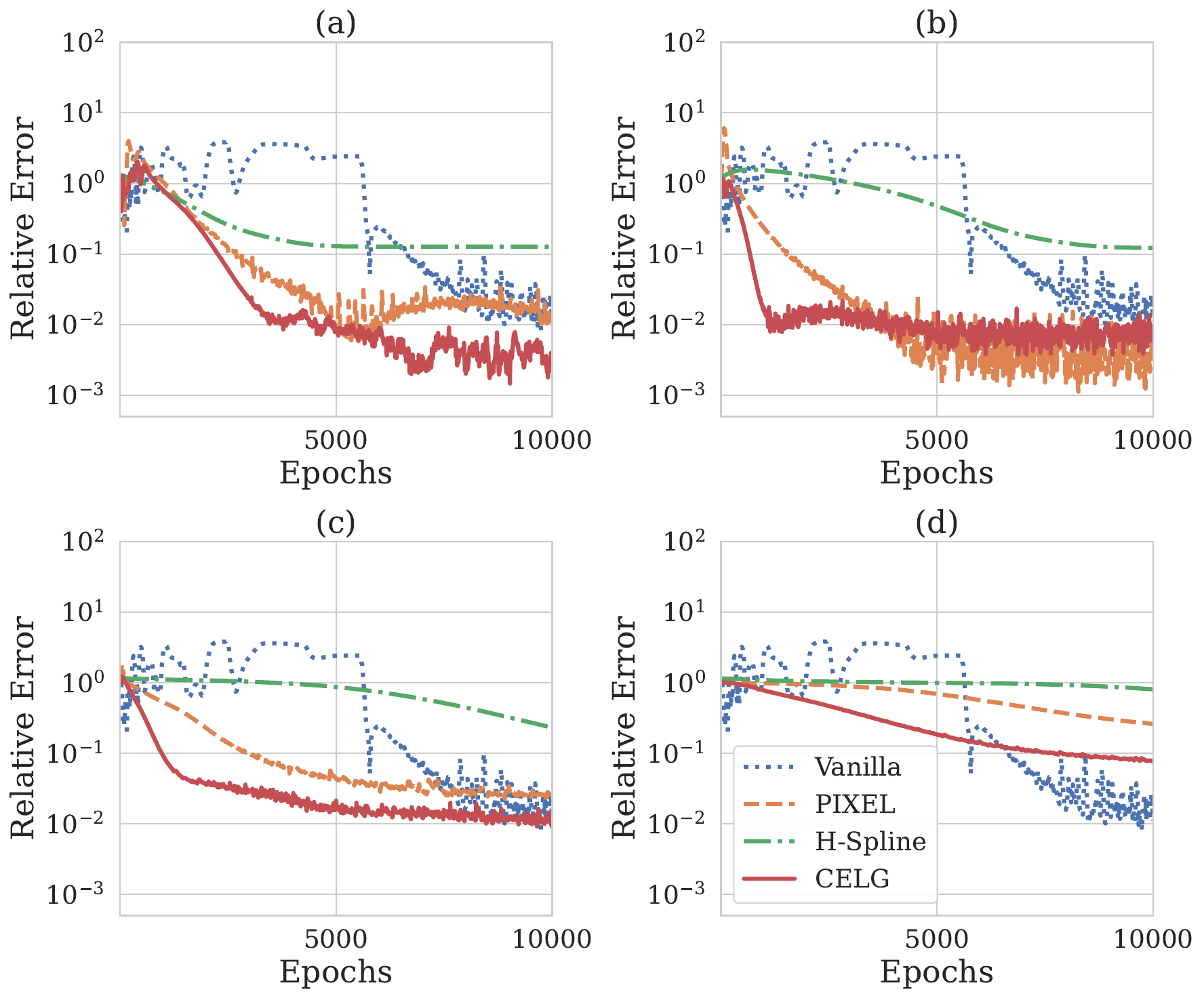} 
\caption{Learning curves for the 1D multi-band Poisson PDE. The grid parameters were set to (a) $R=4$, (b) 8, (c) 16, and (d) 32. }
\label{fig5_Pisson1D_LR}
\end{figure}

\subsection{Training Efficiency}

We compared the training efficiency of the proposed method with that of competitive methods for 1D and 2D multi-band Poisson PDEs with parameters $w_1=1.0$, $w_2=0.1$, $c_1=2$, and $c_2=30$. Fig. \ref{fig3_PissonTrue} shows the analytical solutions of the PDEs, both of which include low- and high-frequency components. The methods compared in this section are the Vanilla PINN, and two grid-based methods (PIXEL~\cite{PXEL_AAAI2023}, and H-Spline~\cite{SplinePINNs_AAAI2022}).

Fig. \ref{fig4_Pisson1D_ResultTrue} shows prediction the results of $u(x)$ for the 1D Poisson equation predicted by these methods at (a) 100, (b) 1,000, (c) 5,000, and (d) 10,000 epochs during training. For all the methods, the loss weights were set to $\lambda_{\mathrm{pde}}=1$ and $\lambda_{\mathrm{bc}}=0.01$. Five hundred and twelve collocation and test points were placed at the regular interval over the domain $\Omega=[0,1]$. The learning ratio of these methods was set to $10^{-3}$. The grid parameter was set to $R=16$. Fig. \ref{fig4_Pisson1D_ResultTrue} shows that the Vanilla method primarily fits the low-frequency components at the early stage of training (Fig. \ref{fig4_Pisson1D_ResultTrue} (a)) and requires many epochs to capture the high-frequency components, as shown by the prediction results at the 10,000 epochs (Fig. \ref{fig4_Pisson1D_ResultTrue} (d)). 
During the intermediate stages (Figs. \ref{fig4_Pisson1D_ResultTrue} (b)--(c)), the predictions differed from the analytical solution; however, it was the transition of the model that likely captured the high-frequency components. The behavior of the Vanilla model demonstrated the spectral bias problem, which caused slow training convergence, which will be discussed in terms of the learning curve (Fig. \ref{fig5_Pisson1D_LR}). In addition, the transient collapse observed in the middle stage is problematic because early stopping of training yields physically meaningless results.
Conversely, the grid-based methods behaved in an opposite manner, with the models first capturing the high-frequency components at an early stage and then gradually capturing the low-frequency components. This result indicates that the grid cell representation facilitates the capture of high-frequency components, which effectively resolves the spectral bias problem. Thus, a stable estimation was achieved without large deviations from the analytical solution during training.

Figs. \ref{fig5_Pisson1D_LR} (a)--(d) show the learning curves of the methods with $R = 4, 8, 16$, and $32$.
The curves of the Vanilla method in these figures are identical to each other because they do not include this parameter.
These figures demonstrate that the grid-based methods converged faster than the Vanilla method.
Among the grid-based methods, H-Spline converged more slowly than PIXEL and CELG owing to non-smooth kernel functions. This was caused by the localized support of the Hermite kernel functions, in which the feature and gradient information are shared only between neighboring cells. Thus, the gradients in model training did not effectively propagate across cell boundaries, resulting in a requirement for a large number of epochs during training for the parameters to converge to their optimal values.
Focusing on the early stage of training, when the number of epochs is smaller than 1,000 epochs, CELG converged faster than PIXEL. This difference was caused by the number of parameters owing to the multigrid structure of PIXEL. Moreover, a comparison of Fig. \ref{fig5_Pisson1D_LR} (d) and Figs. \ref{fig5_Pisson1D_LR} (a)--(c) shows that the excessive number of grid cells degrades training efficiency. 
 
\begin{figure}[p]
\centering
\includegraphics[width=0.50\columnwidth]{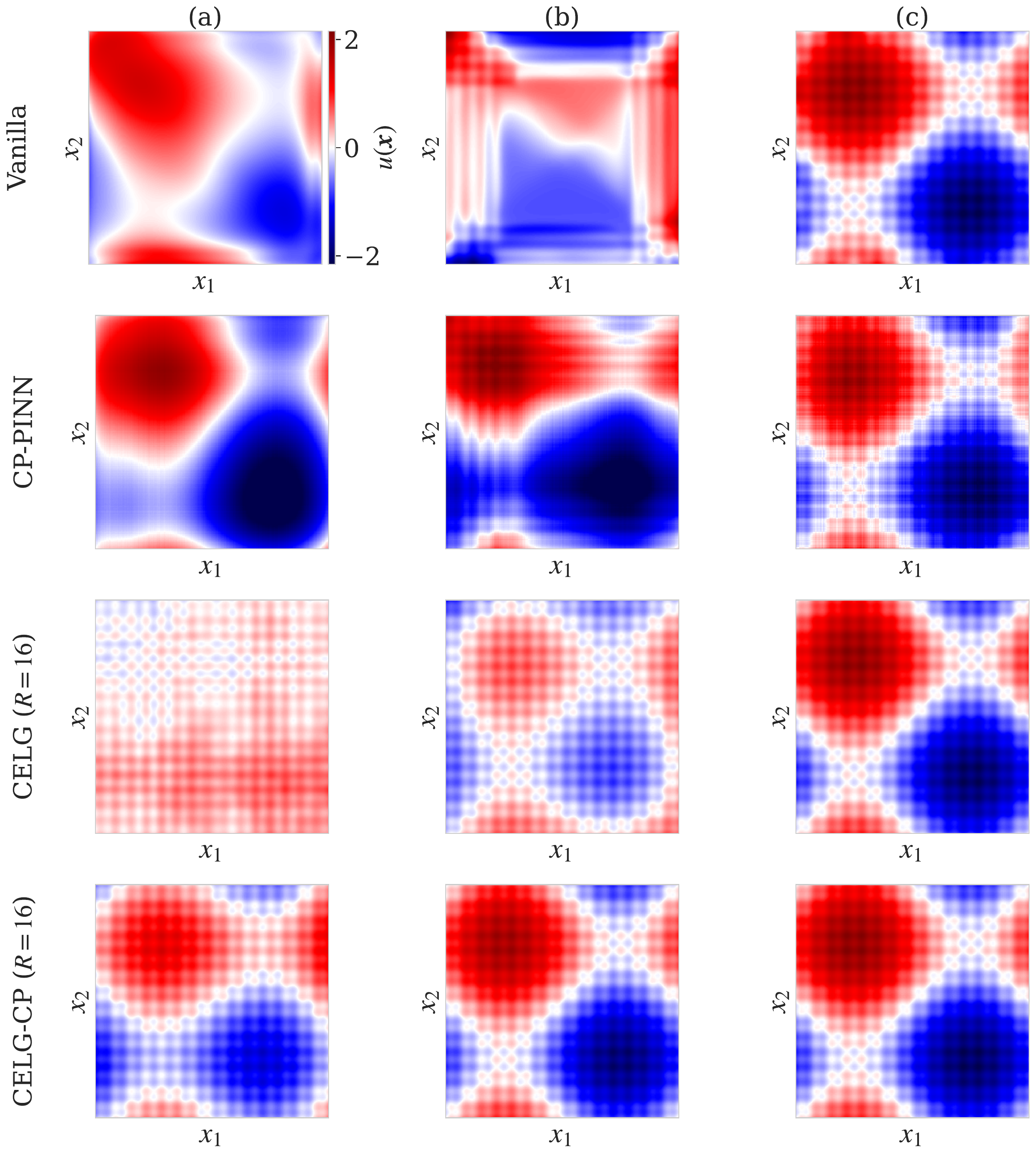} 
\caption{Predictions for the 2D multi-band Poisson PDE by non-grid-based models (Vanilla and CP-PINN) and CELG, CELG-CP ($R = 16$) after (a) 1,000, (b) 5,000, and (c) 50,000 training epochs.}
\label{fig6_Pisson2D_Result}
\vspace{10mm}
\includegraphics[width=0.50\columnwidth]{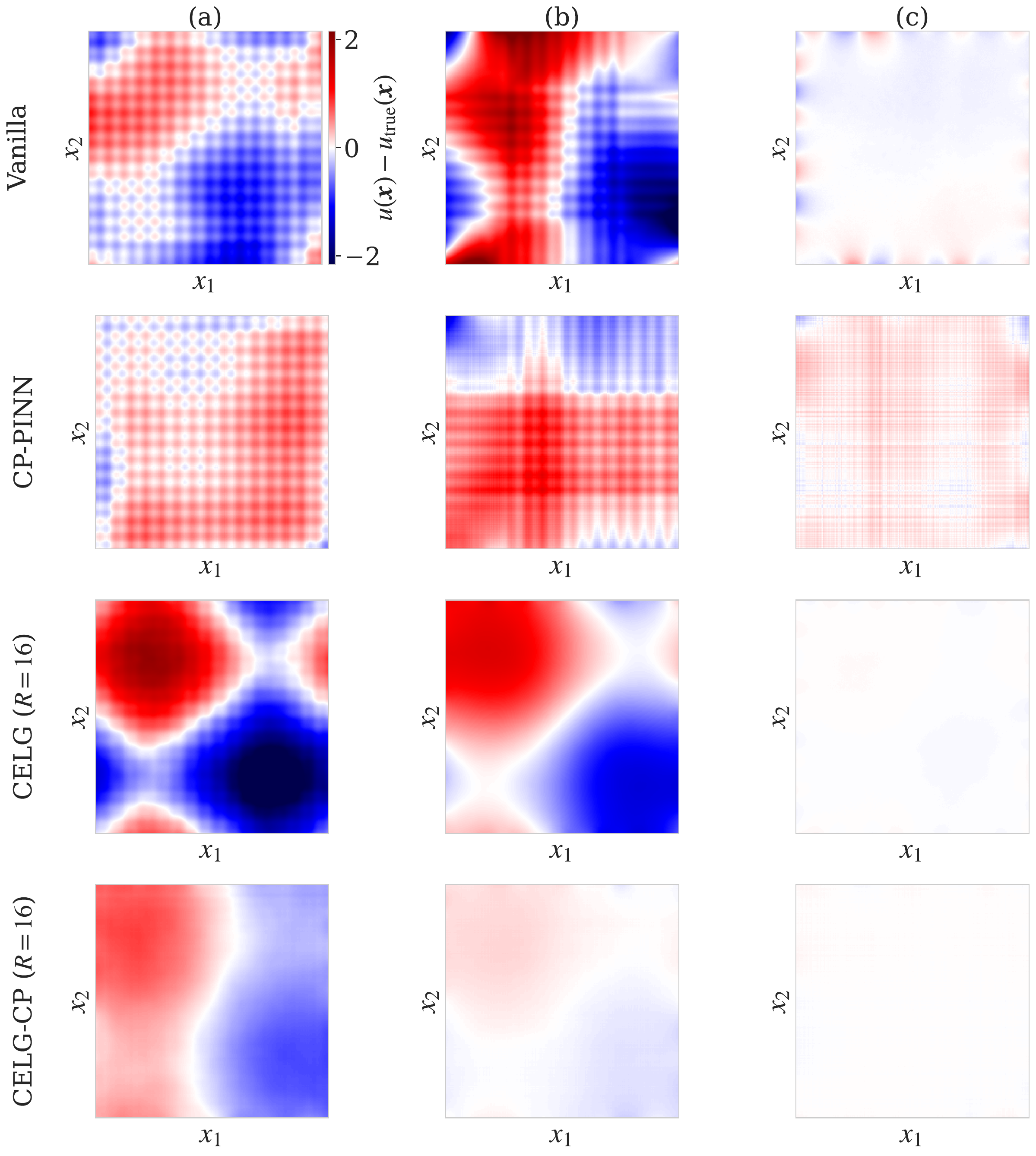} 
\caption{Residual maps for the 2D multi-band Poisson PDE by non-grid-based models (Vanilla and CP-PINN) and CELG, CELG-CP ($R = 16$) after (a) 1,000, (b) 5,000, and (c) 50,000 training epochs.}
\label{fig7_Pisson2D_Residual}
\end{figure}

\begin{figure}[t]
\centering
\includegraphics[width=0.7\columnwidth]{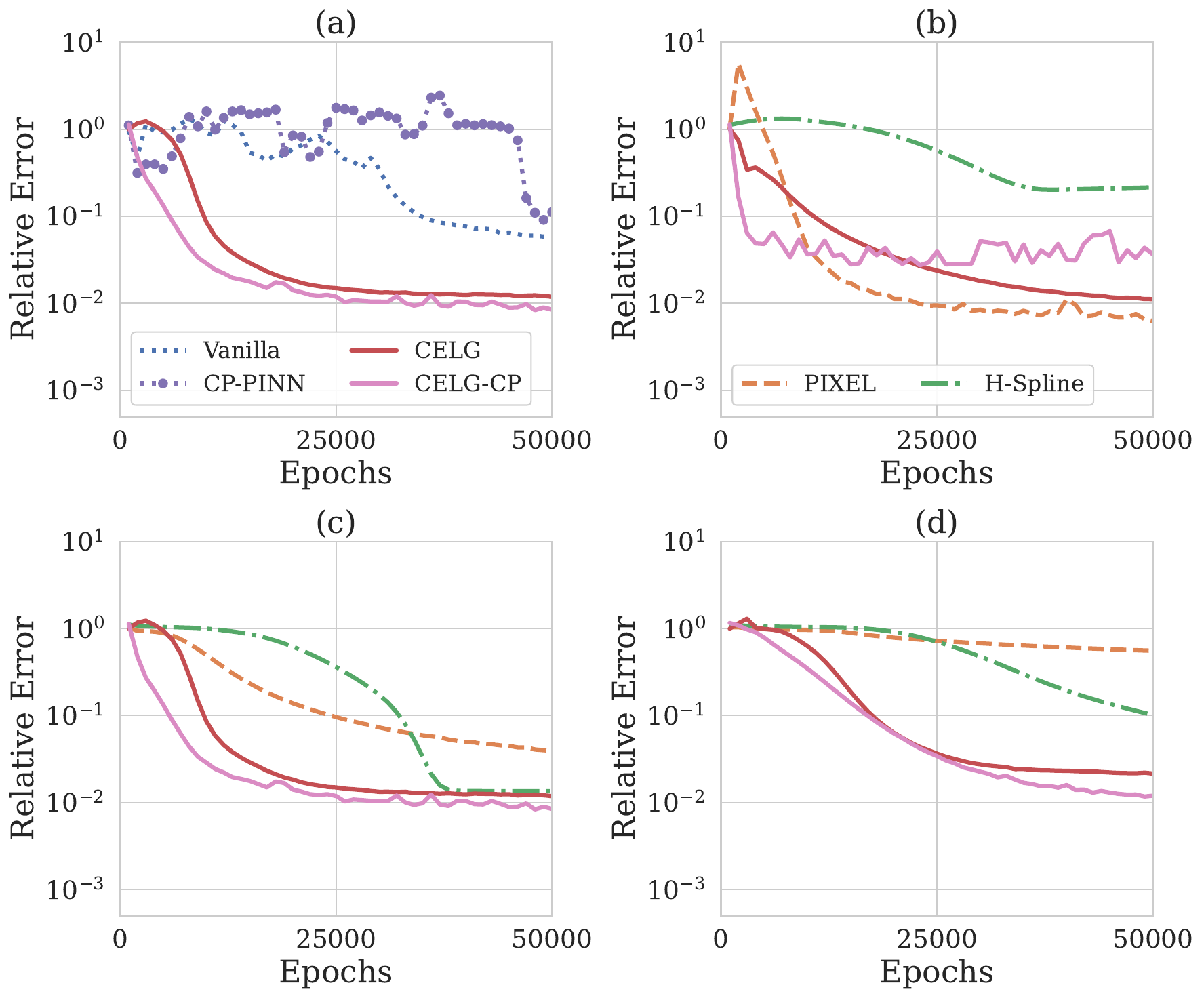} 
\caption{Learning curves for the 2D multi-band Poisson PDE. (a) Curves of CELG and CELG-CP with $R = 16$ and non-grid methods (Vanilla and CP-PINN), and those of grid-based methods with (b) $R = 8$, (c) $R = 16$, and (d) $R = 32$.}
\label{fig8_Pisson2D_LR}
\end{figure}

Fig. \ref{fig6_Pisson2D_Result} shows the function $u(\bm{x})$ of the 2D Poisson equation predicted by CELG and CELG-CP with $R = 16$, vanilla PINN, and CP-PINN during training at (a) $1000$, (b) $5,000$ and (c) $50,000$ epochs. For all methods, the loss weights were set to $\lambda_{\mathrm{pde}}=1$ and $\lambda_{\mathrm{bc}}=0.01$. Two hundred and fifty six collocation and test points were placed at the regular interval over the domain $\Omega=[0,1]^2$. To evaluate the boundary conditions, 1024 points were placed at the regular intervals at the boundary. The learning ratio of these methods was set to $10^{-4}$. Fig. \ref{fig7_Pisson2D_Residual} shows the residual maps between the analytical solution and predictions at the same epochs. CP-PINN is a tensor-factorization-based method that represents the relationships between different spatial-axes, while the coordinates are directly inputted into the model, similar to the vanilla method. CELG-CP is a variant of CELG, and the relationships between different spatial-axes are modeled by tensor factorization. The vanilla and CP-PINN methods captured the low-frequency components during the early training steps, as shown in Fig. \ref{fig6_Pisson2D_Result} (a). 
Subsequently, they gradually captured the high-frequency components in the later steps, as shown in Figs. \ref{fig6_Pisson2D_Result} (b)--(c). Fig. \ref{fig7_Pisson2D_Residual} (c) shows that the residuals of these methods that remained after the training ($50,000$ epochs). These results suggest that non-grid-based methods inherently emphasize low-frequency components, which manifests as spectral bias. Contrarily, the grid-based methods of CELG and CELG-CP exhibit the opposite behaviors, which first capturing the high-frequency components and subsequently adapting to the low-frequency components. As shown in Figs. \ref{fig6_Pisson2D_Result} (b) and \ref{fig7_Pisson2D_Residual}, the speed of convergence to the solution was faster than that in the non-grid methods.

Fig. \ref{fig8_Pisson2D_LR} (a) shows the learning curve of CELG and CELG-CP with $R = 16$, along with those of the non-grid methods of the Vanilla and CP-PINN. CELG and CELG-CP converged much faster than the Vanilla and CP-PINN methods, both of which suffered from the spectral bias. Thus, the spectral bias could be resolved by a grid-based model: however, it could not be resolved by methods that require a direct input of spatial coordinates. Therefore, the problem remains even when a tensor factorization model is used. Figs. \ref{fig8_Pisson2D_LR} (b)--(d) show the learning curves of the grid-based methods for $R = 8$, $16$, and $32$. Similar to the one-dimensional case, CELG and PIXEL outperformed the H-Spline in these three cases. The efficiencies of all methods degraded; however, CELG and CELG-CP were more robust to the setting of $R$ than PIXEL and H-Spline. These observations are consistent with the one-dimensional case. 

\begin{table}[t]
  \centering
  \caption{Training settings for benchmark PDEs. }
  \label{table2_Bench_Setting}
   \setlength{\tabcolsep}{3pt}
   \small
  \begin{tabular}{lcccc}
    \hline
     & Allen--Cahn & Burgers & Flow-Mixing & Helmholtz \\
    \hline
    Domain          & $[0,1]$$\times$$[-1,1]$ & $[0,1]$$\times$$[-1,1]$ & $[0,8]$$\times$$[-4,4]^2$ & $[0,1]^3$ \\
    \#Colloc.    & 50$\times$128 & 50$\times$128 & $64^3$ & $64^3$ \\
    \#IC        & 128 & 128 & $64^2$ & - \\
    \#BC       & $50\times2$ & $50\times2$ & $4\times64^2$ & $6\times64^2$ \\
    $\bm{\lambda} $ & (0.1, 1, 1) & (1, 1, 0.001) & (100, 1, 1) & (1, -, 1) \\
    \#Test           & $100\times256$ & $100\times256$ & $128^3$ & $128^3$ \\
    Epochs                & 200{,}000 & 80{,}000 & 80{,}000 & 50{,}000 \\
    LR         & $10^{-4}$ & $10^{-4}$ & $10^{-3}$ & $10^{-3}$ \\
    \hline
  \end{tabular}
   \setlength{\tabcolsep}{6pt}\\
    Colloc.: collocation, IC: initial condition, BC: boundary condition, \\
    $\bm{\lambda}=(\lambda_{\mathrm{pde}}, \lambda_{\mathrm{init}}, \lambda_{\mathrm{bc}})$, LR: learning rate
\end{table}

\begin{table}[t]
  \begin{center}
  \caption{Number of parameters for benchmark PDEs.} 
  \label{table3_Param_Bench}
  \setlength{\tabcolsep}{3pt}
  \small
  \begin{tabular}{lrrrr}
    \hline
    Method   & $R$   & Allen-Cahn/Burgers  & Flow-Mixing & Helmholtz \\
    \hline
    Vanilla  & --    &  12,737  &   12,801   &    1,985 \\
    CP-PINN  & --    &  25,216  &   37,824   &    5,808 \\
    \hline
    PIXEL      & 8   & 18,753  &  101,697   &  24,433 \\
               & 16  & 45,377  &  637,249   &  158,321 \\
               & 32  & 147,777  &  4,608,321   &  1,151,089 \\
    \hline
    H-Spline   & 8 &   576  &  13,824   &  13,824 \\
               & 16 &   2,304  &  110,592   &  110,592  \\
               & 32 &   9,216  &  884,736   &  884,736 \\
    \hline
    CELG       & 8   &   9,537  &   10,113    &   1,537 \\
               & 16  &  10,561  &   11,649    &   1,921 \\
               & 32 &  12,609  &   14,721    &   2,689 \\
    \hline
    CELG-CP    & 8   &  17,792  &   26,688    &   3,696 \\
               & 16 &  18,816  &   28,224    &   4,080 \\
               & 32 &  20,864  &   31,296    &   4,848 \\
    \hline
  \end{tabular}
  \end{center}
\end{table}

\begin{table}[t]
  \begin{center}
  \caption{Training time (ms./iter.) for benchmark PDEs.}
  \label{table4_Bench_Time}
  \setlength{\tabcolsep}{3pt}
   \small
  \begin{tabular}{lrcccc}
    \hline
    Method  & $R$  & Allen-Cahn           & Burgers             & Flow-Mixing          & Helmholtz \\
    \hline
    Vanilla     & -- & 1.73$\pm$ 0.03         &  1.46$\pm$  0.01       & \bf{1.85$\pm$  0.01}         & 24.73$\pm$ 0.02 \\
    CP-PINN     & -- &  2.04$\pm$ 0.05         &  1.71$\pm$ 0.04        &  1.90$\pm$  0.02         &  4.31$\pm$ 0.23 \\
    \hline
    PIXEL       & 8  &  1.68$\pm$ 0.07         &  2.26$\pm$  0.01       & 34.83$\pm$  0.01         & 24.86$\pm$ 0.01 \\
                & 16 &  1.60$\pm$ 0.07         &  2.07$\pm$  0.01       & 27.57$\pm$  0.01         & 18.77$\pm$ 0.02   \\
                & 32 &  1.63$\pm$ 0.11         &  2.01$\pm$  0.01       & 30.69$\pm$  0.01         & 17.53$\pm$ 0.02   \\
    \hline
    H-Spline    & 8   &  \bf{0.88$\pm$0.13}          &  \bf{0.67$\pm$ 0.04}   & 19.54$\pm$  0.29         & 21.44$\pm$ 0.20 \\
                & 16  &  \bf{0.79$\pm$0.07}     &  \bf{0.70$\pm$ 0.05}   & 17.97$\pm$  0.33         & 19.77$\pm$ 0.48  \\
                & 32  &  \bf{0.84$\pm$0.13}         &  \bf{0.70$\pm$ 0.04}   & 18.16$\pm$  0.30         & 19.88$\pm$ 0.32  \\
    \hline
    CELG      &  8  &  1.70$\pm$ 0.11         &  2.06$\pm$  0.01       & 12.36$\pm$  0.01         & 11.79$\pm$ 0.01 \\
                    & 16 &  1.80$\pm$ 0.11         &  2.28$\pm$  0.01       & 12.62$\pm$  0.01         & 12.11$\pm$ 0.02 \\
                    & 32 &  2.21$\pm$ 0.01         &  2.87$\pm$  0.01       & 13.43$\pm$  0.01         & 12.88$\pm$ 0.01 \\
     \hline
    CELG-CP & 8   &  1.73$\pm$ 0.13         &  1.72$\pm$  0.06       & \bf{1.65$\pm$ 0.02}      & \bf{2.06$\pm$ 0.01} \\
                     & 16 &  1.86$\pm$ 0.14         &  1.84$\pm$  0.06       &  \bf{1.69$\pm$  0.01}         &  \bf{2.18$\pm$ 0.01} \\
                     & 32 &  1.91$\pm$ 0.15         &  2.08$\pm$  0.05       &  2.38$\pm$  0.03         &  \bf{2.50$\pm$ 0.01} \\
    \hline
  \end{tabular}
  \setlength{\tabcolsep}{6pt}
  \end{center}
\end{table}

\begin{figure}[t]
\centering
\includegraphics[width=0.70\columnwidth]{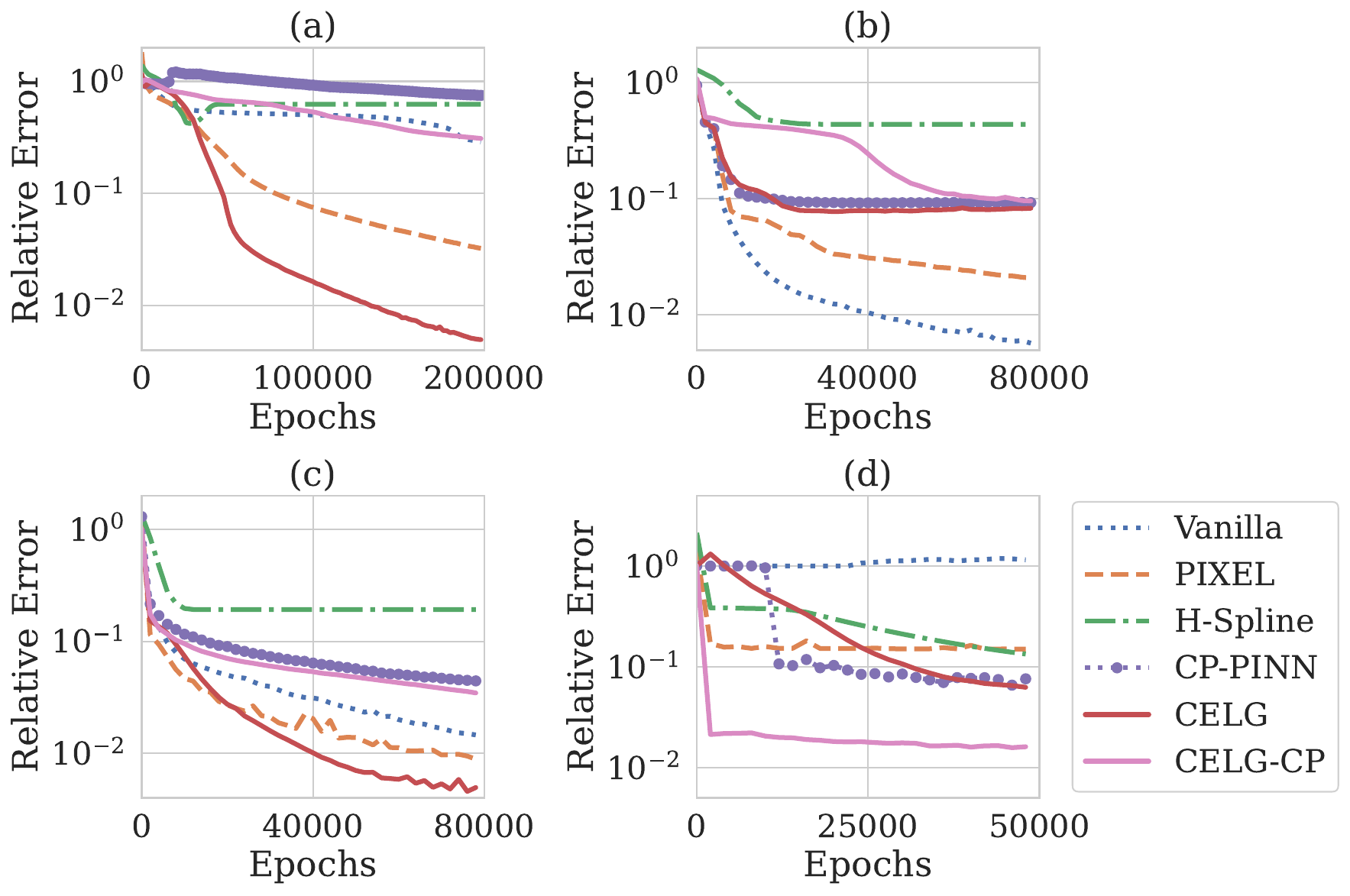} 
\caption{Learning curves for benchmark PDEs: (a) Allen-Cahn, (b) Burgers, (c) Flow-Mixing, and (d) Helmholtz equations.}
\label{fig9_Bench_LR}
\end{figure}

\begin{figure*}[p]
(a) Allen-Cahn\\
\includegraphics[width=0.895\textwidth]{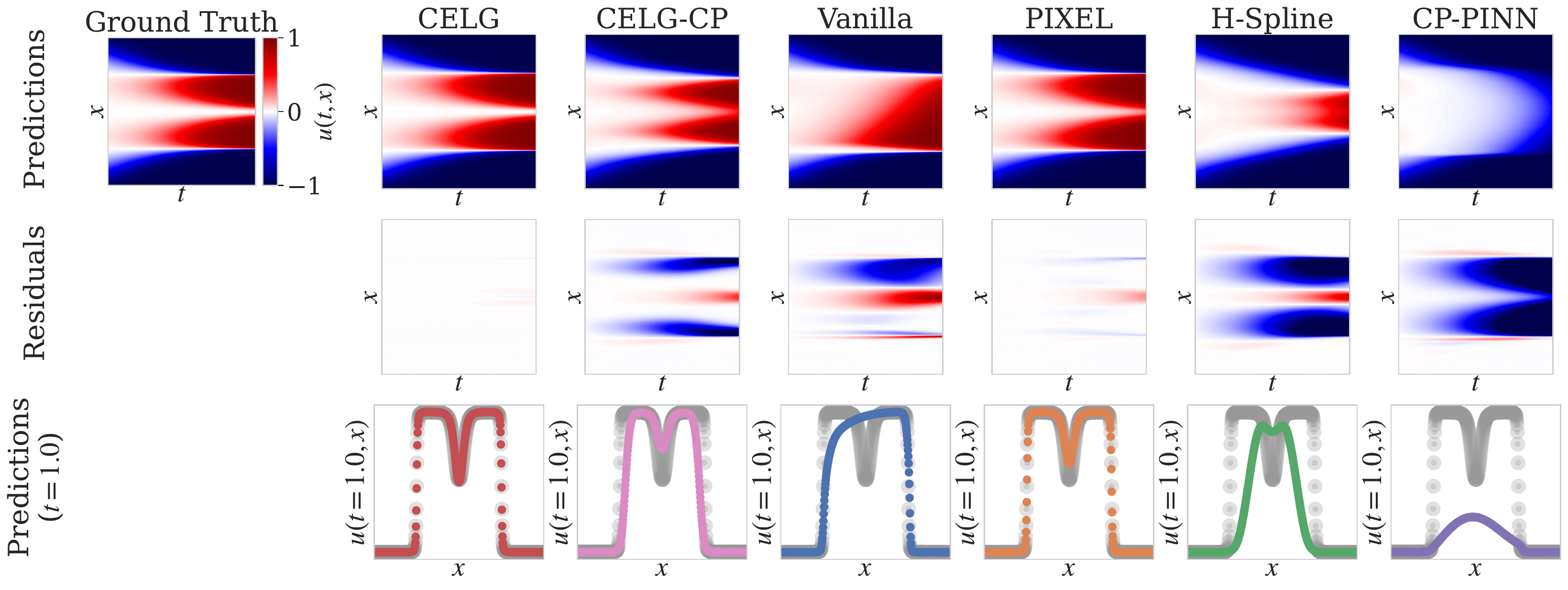} \\
(b) Burgers\\
\includegraphics[width=0.895\textwidth]{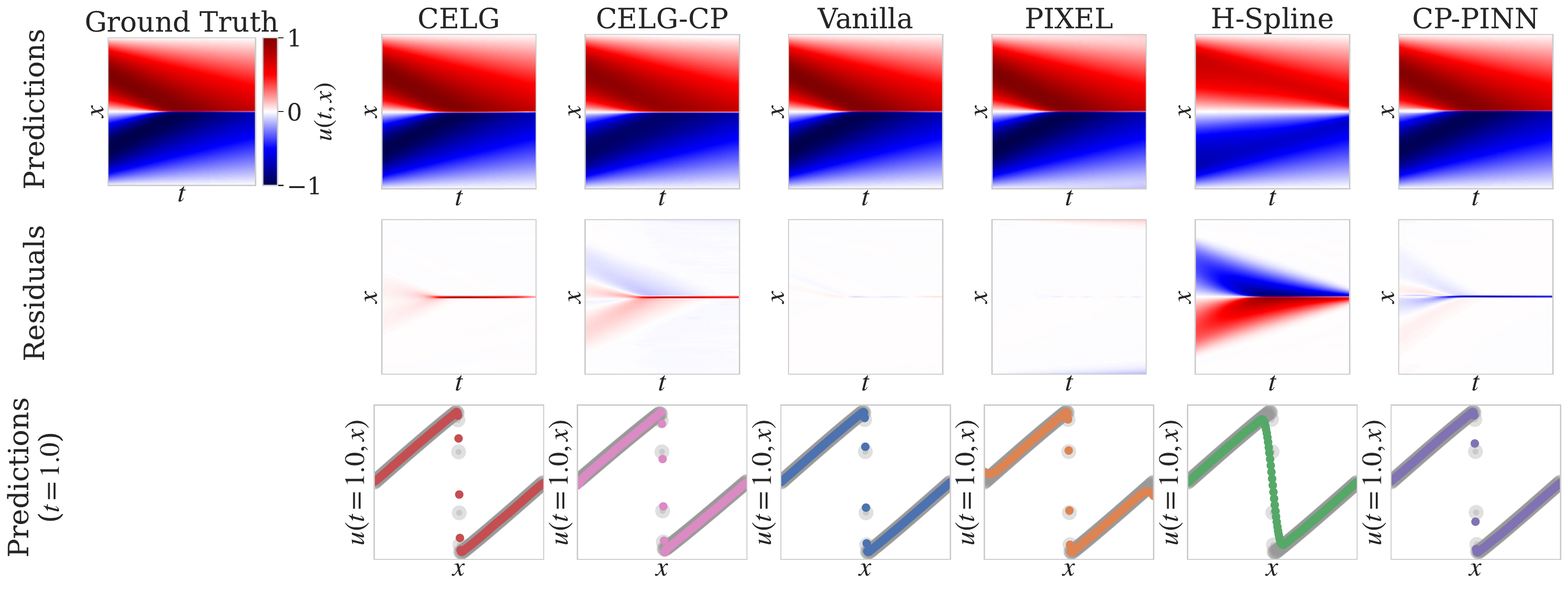} \\
(c) Flow-Mixing\\
\includegraphics[width=0.895\textwidth]{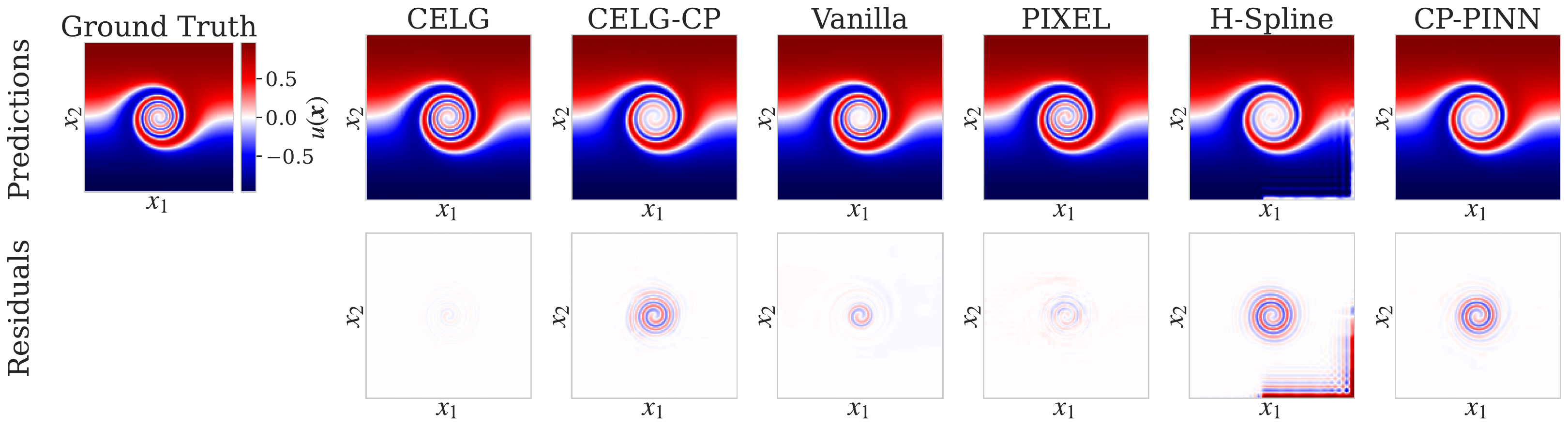} \\
(d) Helmholtz\\
\includegraphics[width=0.895\textwidth]{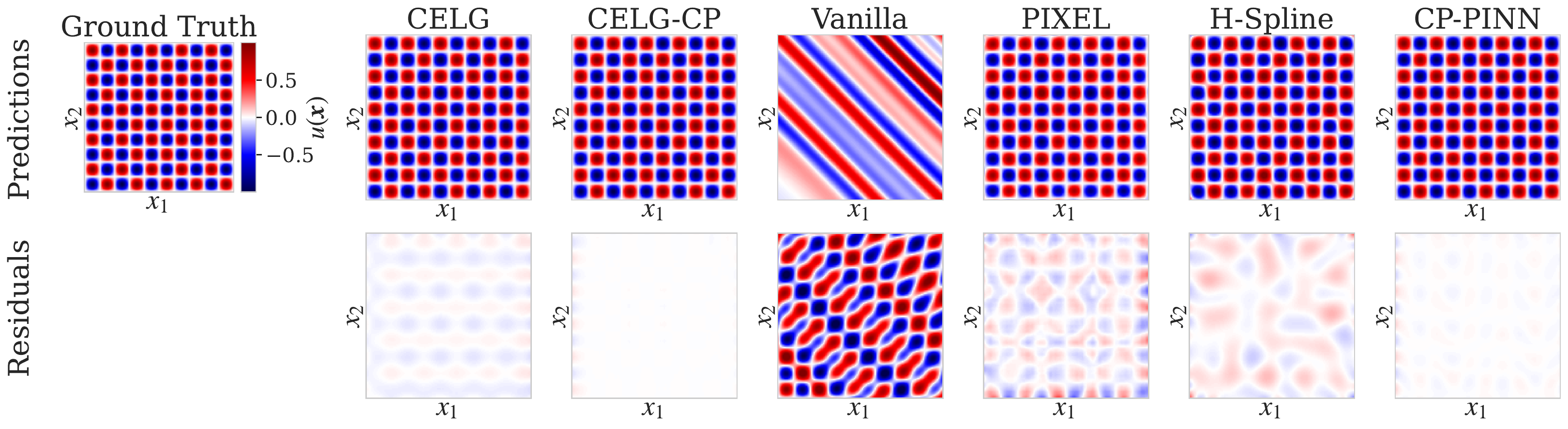}
\caption{Prediction results for (a) Allen-Cahn, (b) Burgers, (c) Flow-Mixing, and (d) Helmholtz equations.
The third rows in (a) and (b) show the prediction results and the analytical solutions (gray) at time $t=1$.
Predictions and residuals at (c) $t=8$ and (d) $x_3=0.75$.}
\label{fig10_Bench_Result}
\end{figure*}

\begin{table}[t]
  \begin{center}
  \caption{Relative errors for benchmark PDEs.}
  \label{table5_Bench_Error}
  \setlength{\tabcolsep}{3pt}
  \small
  \begin{tabular}{lrrrrr}
    \hline
    Method & $R$ & Allen-Cahn & Burgers & Flow-Mixing & Helmholtz \\
    \hline
    Vanilla     & -- & 0.46$\pm$0.23   & \bf{0.08$\pm$0.05}   & 0.02$\pm$0.00  & 1.08$\pm$0.13 \\
    CP-PINN & -- & 0.88$\pm$0.11  & 0.09$\pm$0.00         & 0.05$\pm$0.00   & 0.07$\pm$0.01 \\
    \hline
    PIXEL      & 8    & \bf{0.04$\pm$0.01}  & \bf{0.09$\pm$0.05}   & 0.01$\pm$0.00         & 0.27$\pm$0.03  \\
                    & 16  & 0.62$\pm$0.02         & 0.47$\pm$0.06         & 0.01$\pm$0.00         & 0.14$\pm$0.00  \\
                    & 32  & 0.99$\pm$0.04         & 0.94$\pm$0.04         & 0.05$\pm$0.01         & 0.68$\pm$0.00  \\
    \hline
    H-Spline   & 8   & 0.62$\pm$0.00       & 0.47$\pm$0.02         & 0.18$\pm$0.00         & 0.54$\pm$0.00 \\
               & 16  & 0.15$\pm$0.00       & 0.45$\pm$0.00         & 0.19$\pm$0.00         & 0.12$\pm$0.00  \\
               & 32  & 1.24$\pm$0.01       & 0.91$\pm$0.00         & 0.91$\pm$0.00         & 1.62$\pm$0.00  \\
    \hline
    CELG       & 8   & \bf{0.03$\pm$0.02}  & \bf{0.08$\pm$0.00}    & \bf{0.01$\pm$0.00}    & 0.12$\pm$0.10 \\
               & 16  & \bf{0.14$\pm$0.19}  & 0.11$\pm$0.06         & \bf{0.00$\pm$0.00}    & 0.05$\pm$0.01 \\
               & 32  & 0.98$\pm$0.01       & 0.17$\pm$0.04         & \bf{0.01$\pm$0.00}    & \bf{0.03$\pm$0.02} \\
    \hline
    CELG-CP    & 8   & 0.42$\pm$0.07       & 0.34$\pm$0.31         & 0.05$\pm$0.00         & 0.05$\pm$0.02 \\
               & 16  & 0.27$\pm$0.02       & 0.14$\pm$0.06         & 0.03$\pm$0.00         & \bf{0.01$\pm$0.00} \\
               & 32  & 0.95$\pm$0.04       & 0.35$\pm$0.22         & 0.02$\pm$0.00         & \bf{0.01$\pm$0.00} \\
    \hline
  \end{tabular}
   \setlength{\tabcolsep}{6pt}
  \end{center}
\end{table}

\subsection{Performance for benchmark PDE problems}

We implemented our methods on two- and three-dimensional benchmark PDEs for comparison with existing PINNs. The problems were the Allen-Cahn (2D), Burgers (2D), Flow-Mixing (3D) and the time-independent Helmholtz (3D) equations. The experimental settings are summarized in Table \ref{table2_Bench_Setting}. 
The methods compared in this section are Vanilla and CP-PINN for non-grid-based methods, and PIXEL and H-Spline for grid-based methods. For the Helmholtz equation, the non-grid-based methods employed eight fully connected layers with 16 nodes, whereas the grid-based methods employed four layers with the same number of nodes. For the tensor factorization methods, the same network structure size was assumed for all axes. For the other PDEs, the non-grid-based methods employed four fully connected layers with 64 hidden nodes, whereas the grid-based methods employed two fully connected layers with the same number of nodes. For PIXEL, two grid structures with grid points offset by half-intervals were assumed. The Hermite cubic basis functions were employed for H-Spline. 

Table \ref{table3_Param_Bench} summarizes the number of model parameters. The number of parameters in CELG is significantly smaller than that in PIXEL for all settings. The number of parameters in H-Spline and CELG is comparable for the 2D PDEs, whereas it is significantly lower in CELG for the 3D PDEs. This significant reduction was achieved by using an axis-independent linear grid-cell structure.
Table \ref{table4_Bench_Time} lists the average computation time (ms) per epoch, highlighting the three best averaged values in bold. These values were computed using five results trained using different randomly initialized parameters. Table \ref{table4_Bench_Time} shows that the computational time of H-Spline is comparable to that of the other grid- and MLP-based models (CELG and PIXEL) for the 2D PDEs, whereas CELG is significantly faster than the other methods for the 3D PDEs because of the small number of parameters. Between tensor-based models, CELG-CP is faster than CP-PINNs in most cases.

Fig. \ref{fig9_Bench_LR} shows the learning curves for benchmark PDEs. Our proposed methods (CELG and CELG-CP) and PIXEL achieved relatively better approximations than the other methods for all PDEs. Vanilla and CP-PINN performed worse than these three methods for Allen-Cahn and Helmholtz PDEs, which included high-frequency components. H-Spline performed worse than the other methods for all PDEs because of the non-smooth property of its interpolation kernel discussed in the Appendix A.

Fig. \ref{fig10_Bench_Result} shows the results predicted by trained models. Fig. \ref{fig10_Bench_Result} (a) shows the prediction results for the Allen-Cahn equation. The analytical solution exhibited sharp changes for all values of $t$, which required a high-frequency component for an accurate approximation. This caused training failures of Vanilla and CP-PINNs, as observed in the third row of Fig. \ref{fig10_Bench_Result} (a). Conversely, CELGs and PIXEL yielded accurate approximations. Fig. \ref{fig10_Bench_Result} (b) shows the prediction results for the Burgers equation, in which the analytical solution included a sharp change at only $x=0$. For this problem, most methods, except for H-Spline, captured sharp changes. For all problems, H-Spline performed worse than the other methods owing to the inadequate properties of the interpolation kernel.

Fig. \ref{fig10_Bench_Result} (c) shows the prediction results for the Flow-Mixing problem. The analytical solution indicates that an intense vortex develops over time. The approximation of a function for large $t$ requires high-frequency components. The residual maps in the second row of Fig. \ref{fig10_Bench_Result} (c) show that CELGs successfully approximated the vortex, whereas large residuals around the center remained when Vanilla and CP-PINN were used owing to the spectral bias. Fig. \ref{fig10_Bench_Result} (d) shows the prediction results for the 3D Helmholtz equation. The analytical solution included high-frequency components that could be decomposed along the spatial axes. Vanilla failed the training because of a spectral bias. The results of the grid-based methods were significantly better than those of Vanilla. CELG-CP and CP-PINN outperformed the other methods because their analytical functions did not include nonlinear interactions among the axes.

Table \ref{table5_Bench_Error} summarizes the relative errors. These values were computed by averaging five results trained using different randomly initialized parameters. CELG and CELG-CP achieved the lowest relative errors for all benchmarks except for the Burgers equation. Overall, these results demonstrate that combining a grid structure with cubic spline interpolation enhances the representational power more efficiently than simply increasing the network depth. In addition, the performance of CELG-CP confirms that the proposed framework can flexibly integrate computationally efficient tensor decomposition techniques.

\begin{table}[t]
  \centering
  \caption{Training Settings for High-Dimensional Helmholtz equations.}
  \label{table6_HD_Settings}
  \setlength{\tabcolsep}{3pt}
  \small
  \begin{tabular}{lcccc}
    \hline
    Hyperparameter        & 2D                  & 3D                  & 4D                  & 5D                  \\
    \hline
    Domain          & $[0,1]^2$           & $[0,1]^3$           & $[0,1]^4$           & $[0,1]^5$           \\
    \#Colloc.    & $64^2$              & $64^3$              & $16^4$              & $16^5$              \\
    \#IC        & -                   & -                   & -                   & -                   \\
    \#BC       & $4\times64$         & $6\times64^2$       & $8\times16^3$       & $10\times16^4$      \\
    \#Test           & $64^2$              & $64^3$              & $16^4$              & $16^5$              \\
    Epochs                & 50{,}000            & 80{,}000            & 100{,}000           & 200{,}000           \\
    LR         & $10^{-3}$           & $10^{-3}$           & $10^{-3}$           & $10^{-3}$           \\
    \hline
  \end{tabular}\\
  Colloc.: collocation, IC: initial condition, BC: boundary condition, \\LR: learning rate
\linebreak
  \begin{center}
  \caption{Number of parameters for High-Dimensional Helmholtz equations.}
  \label{table7_HD_Params}
  \setlength{\tabcolsep}{3pt}
  \small
  \begin{tabular}{lrrrrr}
    \hline
    Method & $R$  & 2D          & 3D            & 4D             & 5D         \\
    \hline
    Vanilla & -- &   1,969      &    1,985       &      2,001       &       2,017 \\
    CP-PINN & -- &   3,872      &    5,808       &      7,744       &       9,680 \\
    \hline
    PIXEL &8   &   3,697      &   24,433       &    841,681       &    7,560,145 \\
               &16 &  10,353      &  158,321       &  10,692,561       &  181,743,569 \\
               &32 &  35,953      & 1,151,089       & 151,799,761       & 5,009,332,177 \\
    \hline
    H-Spline & 8   &    576      &   13,824       &    331,776       &    7,962,624 \\
                   & 16 &   2,304      &  110,592       &   5,308,416       &  254,803,968 \\
                   & 32 &   9,216      &  884,736       &  84,934,656       & 8,153,726,976 \\
    \hline
    CELG & 8   &   1,393      &    1,537       &      1,681       &       1,825 \\
               & 16 &   1,649      &    1,921       &      2,193       &       2,465 \\
               & 32 &   2,161      &    2,689       &      3,217       &       3,745 \\
    \hline
    CELG-CP & 8   &   2,464      &    3,696       &      4,928       &       6,160 \\
                     & 16 &   2,720      &    4,080       &      5,440       &       6,800 \\
                     & 32 &   3,232      &    4,848       &      6,464       &       8,080 \\
    \hline
  \end{tabular}
  \end{center}
\end{table}

\subsection{Scalability for High-dimensional PDEs}

We evaluated the scalability of CELG for high-dimensional PDE problems. The PDEs implemented in this experiment were Helmholtz equations  in 2--5 dimensions. Table \ref{table6_HD_Settings} summarizes the experimental settings. The loss weights $\bm{\lambda}$ were set to the same values as in the previous experiment. All methods were implemented once because of their high computational costs. To prevent poor sampling variation owing to the high-dimensional setting, the collocation points were randomly resampled at every epoch. The compared methods were identical to those used in the benchmark experiments. The non-grid-based methods (Vanilla and CP-PINN) assumed eight fully connected networks with 16 hidden nodes, whereas the grid-based methods (CELG, CELG-CP, and PIXEL) assumed four layers with 16 hidden nodes. The other structural settings were the same as those used in the benchmark experiments. Table \ref{table7_HD_Params} summarizes the number of model parameters. The number of parameters in our proposed methods increased moderately with an increase in the number of dimensions, whereas they increased exponentially in PIXEL and H-Spline, owing to their interpolation parameters.

Table \ref{table8_HD_Time} summarizes the average computation times (ms) per epoch. Table \ref{table8_HD_Time} shows that CELG and CELG-CP required less computational time than the other methods. CELG and CELG-CP required less computational time than the other methods. The advantage of using CELG and CELG-CP lies in their superior memory efficiency, allowing the models to fit into the GPU memory even for high-dimensional problems. The other grid-based methods (H-Spline and PIXEL) encountered out-of-memory (OOM) problems. For instance, for a two-dimensional PDE with rank $R = 32$, CELG required approximately one-sixteenth the number of parameters of PIXEL and one-fourth of those of H-Spline (Table \ref{table7_HD_Params}). Under the same conditions in five dimensions, the number of parameters in CELG was less than one-millionth of those of PIXEL and H-Spline, both of which faced OOM errors in the experiments. These observations are consistent with the discussion in Section \ref{sec:complexity}. Table \ref{table9_HD_Error} summarizes the relative errors between the prediction results and the analytical solutions; the three lowest errors are highlighted in bold. Our methods, CELG and CELG-CP, exhibited the lowest relative errors across all dimensions. Overall, these results demonstrate that CELG successfully balances memory efficiency and numerical stability in high-dimensional PDEs.

\begin{table}[t]
  \begin{center}
  \caption{Training Time (ms./iter.) for High-Dimensional Helmholtz equations.}
  \label{table8_HD_Time}
  \setlength{\tabcolsep}{3pt}
  \small
  \begin{tabular}{lccccc}
    \hline
    Method                   &    $R$                           & 2D             & 3D            & 4D            & 5D \\
    \hline
    Vanilla & -- & 13.344      &  43.467       & 56.801        &180.465 \\
    CP-PINN    & -- &  14.421        &  28.907       & 42.602        & 63.602 \\
    \hline
    PIXEL & 8  &  14.276        &  41.424       & 59.750        & OOM \\
     & 16                                      &  14.610        &  38.370       & 63.222        & OOM \\
     & 32                                      &  14.153        &  38.310       & 65.291        & OOM \\
    \hline
    H-Spline & 8  &  11.364        &  36.530       & OOM           & OOM \\
     & 16                                   &  12.182        &  36.128       & OOM           & OOM \\
     & 32                                   &  11.208        &  35.866       & OOM           & OOM \\
    \hline
    CELG & 8                                        &  13.278        &  32.843       & 47.888        & 68.024 \\
     & 16                                       &  14.243        &  34.251       & 49.992        & 83.651 \\
     & 32                                       &  15.846        &  34.140       & 51.398        &108.132 \\
    \hline
    CELG-CP & 8                                     &  12.077        &  24.289       & 40.244        & 72.170 \\
     & 16                                    &  12.457        &  23.377       & 39.176        & 88.153 \\
     & 32                                    &  12.925        &  26.212       & 42.653        &114.139 \\
    \hline
  \end{tabular}
  \end{center}
\end{table}

\begin{table}[t]
  \begin{center}
  \caption{Relative Errors for High-Dimensional Helmholtz equations.}
  \label{table9_HD_Error}
  \setlength{\tabcolsep}{3pt}
  \small
  \begin{tabular}{lrrrrr}
    \hline
    Method   & $R$ & 2D            & 3D            & 4D            & 5D            \\
    \hline
    Vanilla     & -- & 2.03E+00      & 1.80E+00      & 4.29E+00      & 7.36E+00      \\
    CP-PINN & -- & \bf{7.78E-02} & \bf{1.64E-01} & 1.33E+00      & 2.72E+01      \\
    \hline
    PIXEL & 8    & 8.94E-01      & 4.22E-01      & 3.99E+00      & OOM           \\
               & 16  & 9.08E-01      & 9.01E-01      & 1.16E+00      & OOM           \\
               & 32 & 1.02E+00      & 1.07E+00      & 1.01E+00      & OOM           \\
    \hline
    H-Spline & 8   & 1.33E+00      & 2.56E+00      & OOM           & OOM           \\
                   & 16 & 7.46E-01      & 7.25E-01      & OOM           & OOM           \\
                   & 32 & 1.28E+00      & 1.83E+00      & OOM           & OOM           \\
    \hline
    CELG & 8   & 1.08E+00      & 2.87E-01      & 2.12E+00      & 2.60E+01      \\
               & 16 & \bf{1.78E-01} & \bf{1.42E-01} & 3.07E+00      & 3.55E+00      \\
               & 32 & 2.99E-01      & 2.17E-01      & \bf{1.01E+00} & \bf{1.06E+00} \\
    \hline
    CELG-CP & 8   & 3.27E-01      & \bf{9.07E-02} & 1.07E+00      & 3.21E+00      \\
                     & 16 & \bf{1.50E-01} & 2.32E-01      & \bf{3.73E-01} & \bf{1.01E+00} \\
                     & 32 & 2.23E-01      & 1.87E-01      & \bf{9.85E-01} & \bf{1.00E+00} \\
    \hline
  \end{tabular}
  \end{center}
\end{table}

\section{Conclusion}
We proposed a grid-based PINN method that employed grid cells independently along each spatial axis and encoded the spatial coordinates using feature vectors with natural spline interpolation. Numerical experiments for benchmarks and high-dimensional problems demonstrated the effectiveness of the proposed method in improving training convergence, which resolved the spectral bias problem, based on a grid cell with coordination encoding and natural cubic spline functions. Moreover, they demonstrated the efficiency of the proposed method, which outperformed other competitive grid-based PINNs by suppressing the number of parameters.

However, our approach faces three challenges. The first is the generalization of the proposed method to non-rectangular grid cells to improve performance, similar to numerical PDE solvers. To address this problem, efficient and effective interpolation procedures are required. The second is the selection of collocation points. Several studies on efficiently selecting these points have been reported \cite{WU_CMAME2023, Gao_SIAMJSC2023}; however, this is still challenging for high-dimensional problems. The third is to resolve the accuracy degradation around the grid points owing to discontinuities, which was observed in the numerical results for the Burgers equation. We believe that solving these problems will make PINNs more efficient and effective for solving challenging scientific problems.

\section*{Acknowledgements}
This research was partially supported by JST CREST \#JPMJCR2435, and NIMS-Tohoku collaborative partnership grant, Japan.

\appendix
\section{Cosine and Hermite Interpolation kernel}
\label{appendix: Hermite-spline}

\begin{figure}[t]
\centering
\includegraphics[width=0.7\columnwidth]{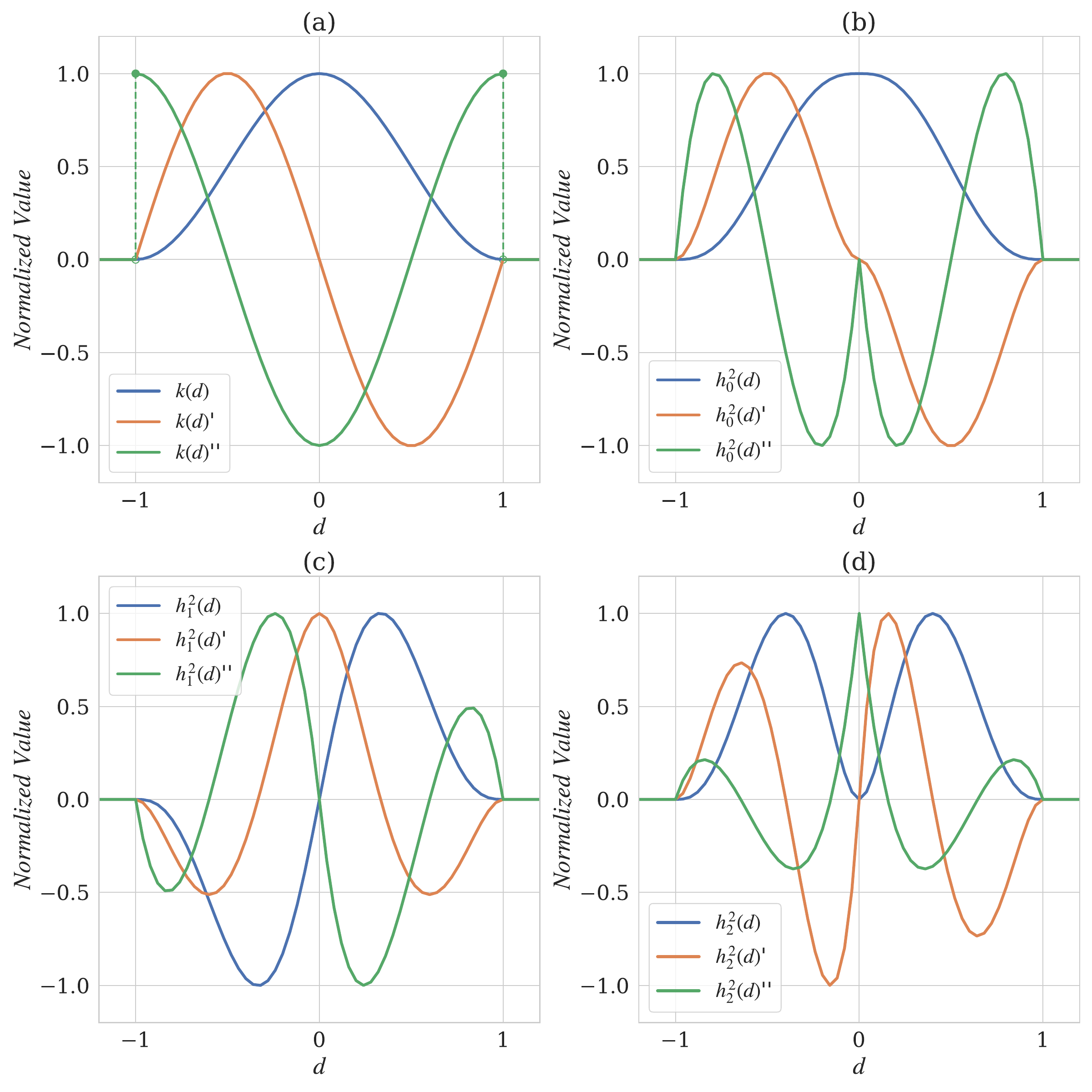} 
\caption{Interpolation kernels and their first and second derivative functions: (a) cosine and (b)--(d) Hermite interpolation kernels.}
\label{fig11_Interp_Kernel}
\end{figure}

This section discusses the continuous properties of the interpolation kernels of the cosine and Hermite interpolations. 
For simplicity, we assume that the grid interval is one. The cosine interpolation kernel~\cite{PXEL_AAAI2023} is defined as a function of the difference between the grid points  $d$:
\begin{eqnarray}
k(d) = \left\{
\begin{array}{cl}
\frac{1}{2}\big\{ 1-\cos \pi (1-|d|) \big\} & \mbox{if}~~ d \in [-1, +1] \\
0 & \mbox{otherwise}
\end{array}
\right.
\end{eqnarray}
The value of the kernel is used as a weight to interpolate the feature vectors.
When the $n$th feature values is $f_n$,
the interpolated feature value at the point $x$ is computed as follows:
\begin{eqnarray}
\hat{f}(x) = \sum_n f_n \cos (\pi (x-x_n)).
\end{eqnarray}
Fig. \ref{fig11_Interp_Kernel} (a) shows the cosine interpolation kernel and the first and second derivative functions.
The original kernel function and the first derivative are smooth, even at the endpoints $d\in\{-1, +1\}$. However, the second derivative is discontinuous at these points. This property may degrade the training performance of PDEs with the second derivatives.

Cubic Hermite interpolation~\cite{SplinePINNs_AAAI2022} is computed as follows:
\begin{eqnarray}
\hat{f}(x) = \sum_n \sum_{i=0}^2 h_i^2 (x-x_n),
\end{eqnarray}
where $h_i^2(\cdot),~i=0,1,2$ are Hermite kernel functions defined by
\begin{eqnarray}
h_0^2(d) &=&\left\{
\begin{array}{ll}
(1-|d|)^3 (1+3|d|+6|d|^2 ) &\\
\hspace{27mm}\mbox{if}~~ d \in [-1, +1] \\
0\hspace{25mm} \mbox{otherwise},
\end{array}
\right.\\
h_1^2(d_n) &=& \left\{
\begin{array}{ll}
2~ \mbox{sign}(d) (1-|d|)^3(|d_n|+3|d|^2 )&\\
\hspace{27mm}\mbox{if}~~ d \in [-1, +1] \\
0\hspace{25mm} \mbox{otherwise},
\end{array}
\right. \\
h_2^2(d) &=& \left\{
\begin{array}{ll}
8 (1-|d|)^3 |d|^2 & \mbox{if}~~ d \in [-1, +1] \\
0 & \mbox{otherwise},
\end{array}
\right.
\end{eqnarray}
where $\mbox{sign}(\cdot)$ is +1 for positive values; otherwise, it is -1. Figs. \ref{fig11_Interp_Kernel} (b)--(c) show Hermite interpolation kernel functions and their first and second derivative functions. In contrast to the cosine interpolation, a Hermite kernels and their derivatives are continuous at the end points. However, not all second derivatives are smooth at the end points, for example, those of $h_0^2(d)$ and $h_0^2(d)$ at grid points ($d=0$). Similar to that for the cosine kernel, this property may degrade the training performance. Figs. \ref{fig11_Interp_Kernel} (b)--(c) shows that Hermite kernels and their derivatives shrink to zero at the endpoints, which makes it difficult for the neighboring grid points to share feature information. This property slows down the training convergence, as demonstrated by our numerical experiments.

\bibliographystyle{unsrtnat}
\bibliography{celg}  

\end{document}